%% file: travelAgent.tex
\documentclass{article}
\usepackage{arxiv}
\usepackage[utf8]{inputenc} 
\usepackage[T1]{fontenc}    
\usepackage{hyperref}       
\usepackage{url}            
\usepackage{booktabs}       
\usepackage{amsfonts}       
\usepackage{nicefrac}       
\usepackage{microtype}      
\usepackage{cleveref}       
\usepackage{graphicx}
\usepackage{natbib}
\usepackage{doi}
\usepackage{verbatim}
\usepackage{float}

\usepackage{longtable}
\usepackage{array}

\title{TravelAgent: Generative Agents in the Built Environment}
\author{
    \href{https://orcid.org/0000-0002-5125-1059}{\includegraphics[scale=0.015]{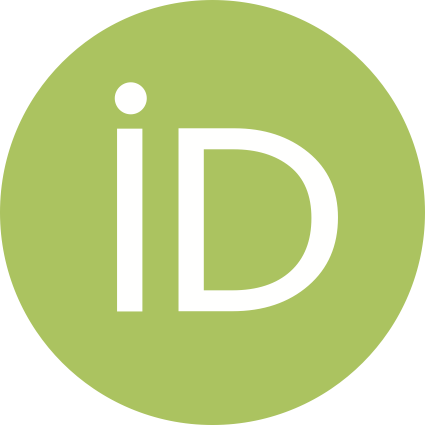}\hspace{1mm}Ariel Noyman\thanks{These authors contributed equally to this work.}}
    \\
    MIT Media Lab
    \\
    \texttt{noyman@mit.edu}
    \\
    \and
    \href{https://orcid.org/0009-0002-1591-4131}{\includegraphics[scale=0.015]{figs/orcid.png}
    \hspace{1mm}\textbf{Kai Hu}\textsuperscript{*}}
    \\
    South China University of Technology
    \\
    \texttt{arhukai@mail.scut.edu.cn}
    \\
    \and
    \hspace{1mm}\textbf{Kent Larson}
    \\
    MIT Media Lab
    \\
    \texttt{kll@mit.edu}
    \\
}

\begin{document}
\maketitle


\begin{abstract}
    Understanding human behavior in built environments is critical for designing functional, user-centered urban spaces. Traditional approaches—such as manual observations, surveys, and simplified simulations—often fail to capture the complexity and dynamics of real-world behavior. To address these limitations, we introduce \textbf{TravelAgent}, a novel simulation platform that models pedestrian navigation and activity patterns across diverse indoor and outdoor environments under varying contextual and environmental conditions. TravelAgent leverages generative agents integrated into 3D virtual environments, enabling agents to process multimodal sensory inputs and exhibit human-like decision-making, behavior, and adaptation. Through experiments, including navigation, wayfinding, and free exploration, we analyze data from 100 simulations comprising 1,898 agent steps across diverse spatial layouts and agent archetypes, achieving an overall task completion rate of 76\%. Using spatial, linguistic, and sentiment analyses, we show how agents perceive, adapt to, or struggle with their surroundings and assigned tasks. Our findings highlight the potential of TravelAgent as a tool for urban design, spatial cognition research, and agent-based modeling. We discuss key challenges and opportunities in deploying generative agents for the evaluation and refinement of spatial designs, proposing TravelAgent as a new paradigm for simulating and understanding human experiences in built environments.
\end{abstract}




\section{Introduction}

\subsection{Background}
Human behavior in urban environments is central to the design of spaces that are functional, inclusive, and responsive to the diverse needs of their users~\citep{gehl2010,jacobs1961death,whyte1980}. These behaviors shaped by a complex interplay of spatial configurations, visual stimuli, social dynamics, and individual preferences~\citep{whyte1980}. This wide range of factors makes it challenging to asses how people will navigate, interact, and experience urban spaces, particularly in early design stages~\citep{hillier1984,batty2000visualizing}.
\newline
Throughout the design process, architects, urban planners, city officials, and stakeholders often rely on a variety of methods to evaluate the impact of design decisions on human behavior~\citep{gehl2010,batty2022digital}. These include manual observations, surveys, focus groups, and computational simulations and visualizations~\citep{noyman2022cityscope,banerjee2011companion}. While these methods offer valuable insights, they often fall short of capturing the full complexity of human experience in real-world settings~\citep{perez2016limitations}. Manual observations are time-consuming and limited in scope, and surveys and focus groups may not fully represent the diversity of user experiences and their particular preferences~\citep{epstein1999agent}. Computational simulations, such as Agent Based Models (ABMs), provide a more scalable and controlled environment for studying human behavior, but often rely on simplified heuristics that do not fully capture the richness of human interactions~\citep{doi:10.1177/23998083211051139,chopraLimitsAgencyAgentbased2024,perez2016limitations}.
\newline
Recent advances in generative models created new possibilities for simulating human-like behavior in urban environments~\citep{luca2021survey}. Large Language Models (LLMs) and Generative Agents (GAs) have demonstrated the ability to perform more complex tasks in response to natural language prompts~\citep{park2023generative}. The primary advantage of generative agents lies in their ability to perform well in non-deterministic scenarios, where a clear set of rules may not exist~\citep{park2023generative,kaiya2023lyfeagentsgenerativeagents}. These systems can simulate multi-step tasks by chaining together a series of actions guided by natural-language prompts~\citep{park2023generative}.

\subsection{TravelAgent: Generative Agents Simulation in Urban Environments}
Building upon recent advancements in generative models, we aim to bridge the gap between urban simulations, ABM, and cognitive analysis of built spaces. Here we introduce \textbf{TravelAgent} (TA), a simulation platform that leverages Generative Agents (GA) to model and analyze human-like behavior and experiences in built environments. TravelAgent enables computational agents to explore, navigate, observe, and interact with their surroundings from a human perspective. Unlike traditional simulations that rely on predefined logic or pre-trained behavioral models~\citep{chopraLimitsAgencyAgentbased2024}, TA incorporates cognitive principles such as spatial memory, visual perception, and context-based reasoning to facilitate autonomous decision-making~\citep{weiChainofThoughtPromptingElicits2023}. We present the TA platform, a web-based tool designed to allow spatial designers to easily simulate pedestrian navigation tasks or open-ended exploratory activities, in various interior and exterior spaces under different environmental conditions.
\newline
TravelAgent contributes to the field of urban simulation in the following ways: (i) It introduces an accessible interface for simulating behaviors ranging from routine navigation tasks to open-ended exploratory activities, and applicable across scales, from early-stage, schematic urban-planning to detailed, room-scale interior layouts. (ii) It provides data and analysis of 100 simulations with 1,898 steps,ranging across various settings, environments, and agent archetypes, achieving a task completion rate of approximately 75\%. (iii) It offers a comprehensive analysis of the agents' behaviors, decision-making processes, and adaptability to unforeseen circumstances, and proposes a novel approach to evaluate and refine spatial designs based on human-like behaviors and experiences.

\subsection{Terminology}
Agent-Based Models (ABMs) are computational models that simulate the actions and interactions of agents in a given computational environment~\citep{bonabeauAgentbasedModelingMethods2002}. ABMs are widely used in urban simulation to study complex systems, such as traffic flow, pedestrian dynamics, or infrastructural efficiency~\citep{CHEN2012166,batty2022digital}. Generative Agents (GA) are a new class of Large Language Models (LLM), that preform multi-step tasks in response to natural language prompts~\citep{weiChainofThoughtPromptingElicits2023,park2023generative}.
In this study, `TravelAgent' (TA), refers to a platform combining Generative Agents (GA) with Agent-Based Modeling (ABM) to simulate agent behavior in built environments. Here, `agent' denotes the simulated entity of an eye-level computational actor in 3D environments, while `simulation' pertains to the process of generating and observing the agent behavior. The platform leverages `sensory inputs' which are different data steams perceived by agents to interact with their surroundings, and `cognitive streams', the structured representations of agents' thoughts, observations, and actions throughout the simulations.

\subsection{Paper Organization}
The remainder of this paper is organized as follows: Section \ref{sec_related_work} provides an overview of related work in spatial cognition, agent-based modeling, and pedestrian experience in urban design. Section \ref{sec_methodology} describes the methodology and implementation of the TravelAgent platform, including agent simulation, sensory inputs, and data collection. Section \ref{sec_case_studies} presents several case studies of TravelAgent in different environments. Section \ref{sec_results} discusses the results and analysis of the case studies, and Section \ref{sec_discussion} concludes with a summary of the contributions, potential impact, and future directions of TravelAgent.

\section{Related Work} \label{sec_related_work}
Understanding human behavior in urban environments is a complex challenge that intersects multiple disciplines, including urban planning, spatial cognition, and computational modeling. This section reviews the key areas of research that inform the development of the TA platform, particularly focusing on the integration of computational models in simulating human behavior in the built environment.

\subsection{Human Experience in Urban Design and City Planning}
The importance of human-centered design principles, such as walkability, accessibility, and public space quality, has been widely recognized in shaping successful built environments~\citep{jacobs1961death,gehl2010,lynch1960image,banerjee2011companion,ameli2015dobetter}. The integration of new technologies has enabled designers to better evaluate urban interventions in immersive ways, enhancing design processes, stakeholder engagement, and decision-making~\citep{batty2000visualizing,billger2017search,yin2017streetlevel,shen2018streetvizor}. Computerized models, simulations, and visualizations can clarify the outcomes of design choices, including building design and spatial configurations~\citep{smith1998visual,batty2000visualizing}. Nevertheless, these digital design aids cannot fully embrace the nature of human behavior in built environments. Often, these tools are only accessible to few experts and fewer stakeholders; even when available, they tend to represent a highly-curated vision of the design, lacking actual interaction and experience~\citep{lin2013cognition,tabrizian2020viewscape,banerjee2011companion}. Newer technologies, such as virtual, augmented, or mixed reality (VR, AR, MR), have the potential to bridge this gap by providing more immersive and interactive experiences for urban design and planning~\citep{batty2022digital}; however, these tools are often limited to a single user or small group, and require specialized equipment and expertise~\citep{fonseca2016informal}, making them less accessible for widespread use in urban design and planning. In that context, the problem of simulating human experience in urban environments remains an open challenge, as traditional methods are often limited in scope, scale, and realism~\citep{batty2022digital,banerjee2011companion,gath2024role}.

\subsection{Spatial Cognition and Human Navigation}
Spatial cognition studies how humans perceive, interpret, and navigate physical spaces~\citep{montello1993scale,golledge1999human}. Research in this area emphasizes the importance of environmental cues, such as landmarks, signage, and spatial configurations, in facilitating wayfinding and orientation~\citep{javadi2017hippocampal,epstein2017cognitive,gehl2010,jacobs1961death,lynch1960image}. Empirical studies have demonstrated that humans rely on a combination of egocentric (self-to-object) and allocentric (object-to-object) representations when navigating unfamiliar environments~\citep{burgess2006spatial,klatzky2008allocentric,mou2004roles}. Cognitive maps-mental representations of physical spaces- are constructed through experience and are essential for efficient navigation and spatial memory~\citep{burgess2017cognitive,epstein2017cognitive}. Advances in neuroimaging and computational modeling have provided new insights into the mechanisms underlying spatial cognition, memory, and navigation~\citep{epstein2017cognitive}. Controlled settings, including the usage of immersive technologies, have also been employed to study spatial cognition, providing valuable data for design and planning~\citep{batty2022digital,gath2024role}. Incorporating principles of spatial cognition into computational models can enhance the realism of simulated agents, allowing them to exhibit human-like navigation behaviors~\citep{milivojevic2013mnemonic,epstein1999agent}.

\subsection{Agent-Based Modeling in Urban Simulation}
Agent-Based Modeling (ABM) has been instrumental in urban simulation, enabling researchers to model the actions and interactions of agents within controlled environments~\citep{bankesAgentbasedModelingRevolution2002,batty2022digital}. ABMs have been applied to a wide range of urban studies, including traffic flow analysis~\citep{nguyenOverviewAgentbasedTraffic2021,grignard2018cityscope}, pedestrian dynamics~\citep{filomenaModellingEffectLandmarks2021}, and urban policy evaluation~\citep{perez2016limitations}. However, traditional ABMs often rely on deterministic heuristics and predefined behavioral rules~\citep{railsback2019agent}, limiting their ability to capture the complexity and variability of human behavior~\citep{zhang2024generative,epstein1999agent}.
Recent research suggested integrating ABMs with machine learning techniques, such as Reinforcement Learning (RL) and Neural Networks (NN)~\citep{kobayashi2023modeling,li2020agent}, to model more adaptive behaviors. Yet, these approaches typically require extensive data, training, and validation and can be computationally intensive~\citep{chopraLimitsAgencyAgentbased2024}. Even when combined with NN, the challenge remains in simulating agents that performs beyond their intrinsic heuristics or learnt distribution; These models will most likely fail to response to unpredictable and complex environmental stimuli~\citep{chen2019touchdown}.

\subsection{Generative Agents (GA)}
The emergence of Large Language Models (LLMs) has opened new avenues for simulating human-like behavior through natural language and generative agents~\citep{park2023generative}. These models process natural language inputs and generate contextually relevant responses, enabling more sophisticated interactions between agents and their environments~\citep{chopraLimitsAgencyAgentbased2024}. By integrating GA with perception modules such as computer vision, agents can interpret and respond to complex environmental stimuli~\citep{reid2024gemini}. In the context of urban simulation, combining LLMs with agent-based models enables agents that can navigate and interact with physical environments~\citep{yang2025v}. These agents can process visual inputs, maintain spatial memory, and make decisions based on prior knowledge and real-time observations, while flexibly adapting to changing conditions and unforeseen events~\citep{park2023generative}.
\newline
GA architectures typically include \textit{Memory Stream} to record their experiences; \textit{Reflection} mechanism to synthesize memories into higher-level concepts; and \textit{Planning} to translate reflections into action plans~\citep{huangUnderstandingPlanningLLM2024,yaoReActSynergizingReasoning2023,zhang2024generative}. Reasoning techniques, such as Chain-of-Thought (CoT) and Tree-of-Thought (ToT), allow GA to decompose tasks into manageable steps and adapt dynamically based on environmental feedback~\citep{weiChainofThoughtPromptingElicits2023,yaoTreeThoughtsDeliberate}. These compound systems can independently tackle more complex tasks, acting collaboratively rather than relying solely on predefined rules or initial prompting~\citep{weiChainofThoughtPromptingElicits2023,yaoTreeThoughtsDeliberate}.
\newline
In urban simulation, GA may offer a promising approach to model more realistic human behaviors. Unlike traditional ABMs, which rely on predefined rules and heuristics, GA can adapt to new environments, tasks, and stimuli, providing more nuanced and context-aware responses~\citep{park2023generative}. By incorporating open-ended decision-making processes, GA can provide insights into how individuals might interact with their environment in ways that traditional ABMs cannot~\citep{ghaffarzadegan2023generative}. This integration has the potential to enhance the design and evaluation of urban spaces by providing more human-centric insight.
\newline
Despite their promise, GA face several challenges. Efficient memory management is critical to avoid overload while maintaining relevant information and avoiding hallucinations~\citep{hatalisMemoryMattersNeed2024}. Mimicking human processes such as reflection and experience remains a challenge due to current limitations of LLMs in understanding context and nuance~\citep{mirzadehGSMSymbolicUnderstandingLimitations2024}. Additionally, the quality of GA reasoning mechanisms is intertwined with their initiation and orchestration, making them sensitive to minor changes in setup and initialization~\citep{zhang2024generative}.

\begin{figure}
    \centering
    \includegraphics[width=1\textwidth]{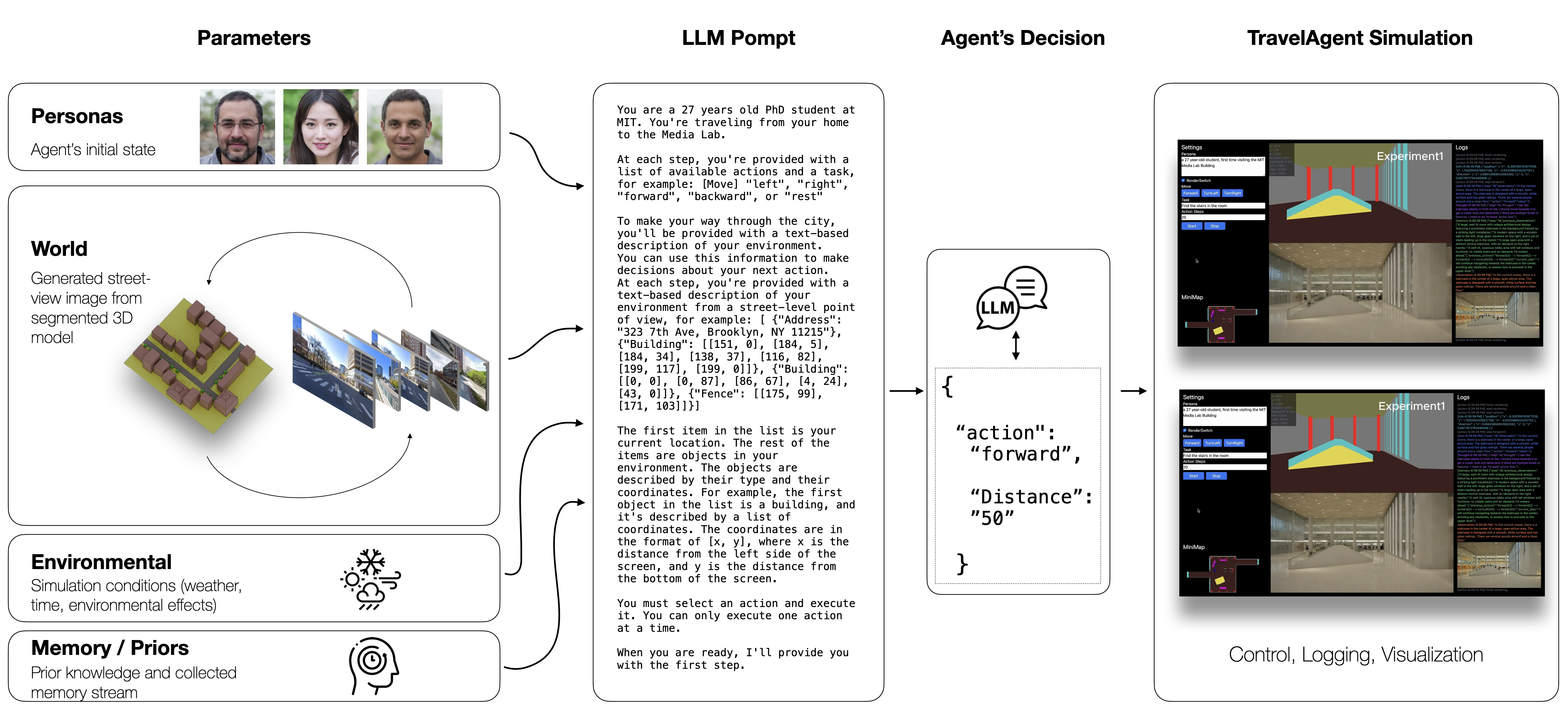}
    \caption{A schematic representation of the TravelAgent system. (left) TAs are initialized with various parameters defining the agent persona and its environment. (center) At each step, the agent employs Chain-of-Thought (CoT) to process sensory inputs, plan actions, and make decisions. (right) The agent executes actions within the environment and updates its internal memory based on its experiences.}
    \label{fig:ta-scheme}
\end{figure}


\section{Methodology} \label{sec_methodology}

\subsection{Agent Simulation}

\begin{figure}
    \centering
    \includegraphics[width=1\textwidth]{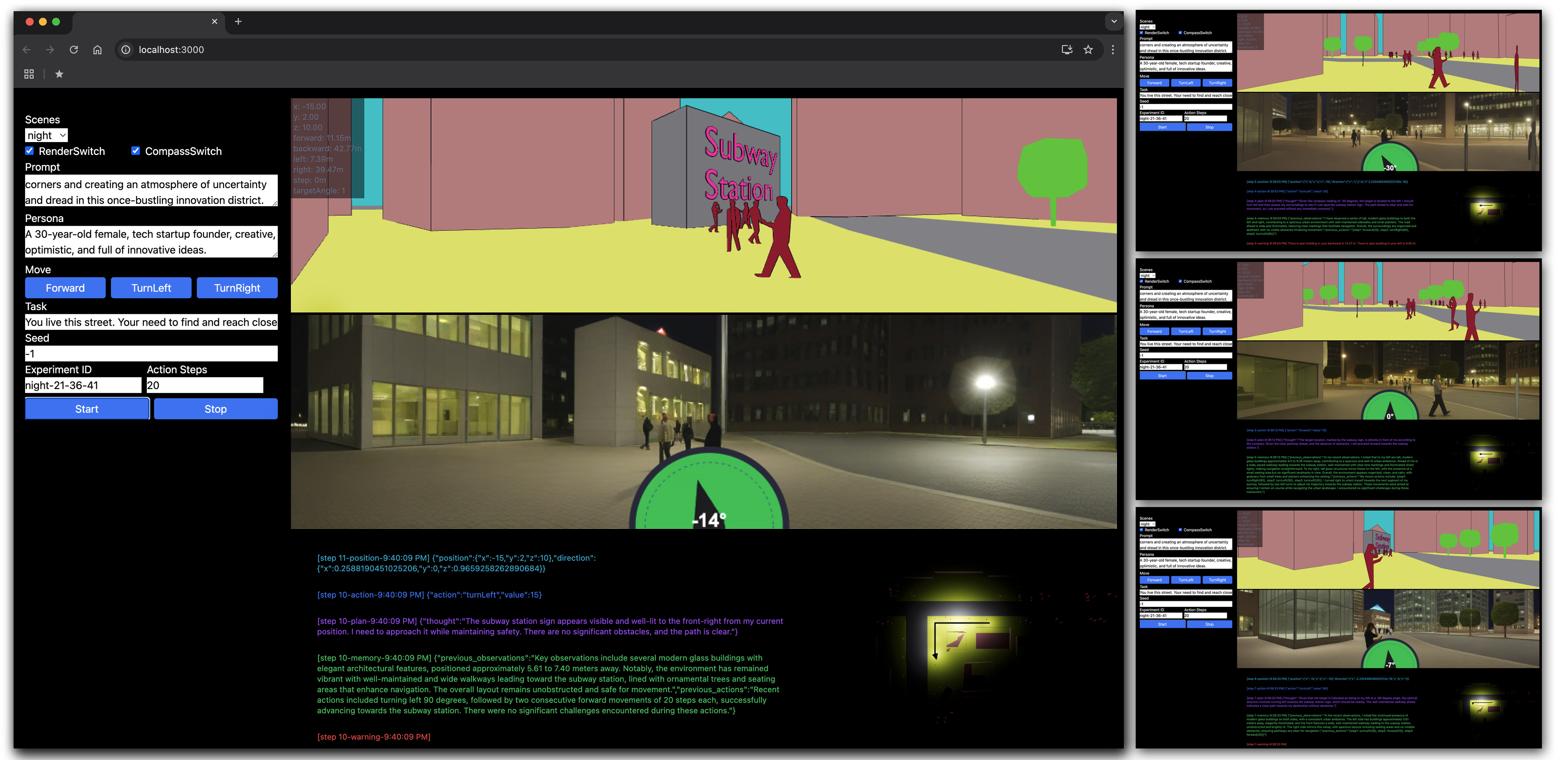}
    \caption{TravelAgent interface. The web app provides an end-to-end experimentation environment for testing and evaluating TravelAgents. (left) Initial settings and inputs provided to the agent; (bottom) Output log of the Chain-of-Thought process: [orange] are the agent's observations, [green] are the agent's memories, [purple] are the agent's plans, [blue] are the actions/decisions. (top) Panoramic street-level view of the environment. The rudimentary 3D environment is guiding an SDXL image generation model to create eye-level images, as well as to provide depth estimation, and collision information. Simulating in generative environments allows users to change scenarios, agents' profiles, and tasks by simply updating a short textual prompt, as shown on the right column.}
    \label{fig:ta-platform}
\end{figure}

\begin{figure}
    \centering
    \includegraphics[width=1\textwidth]{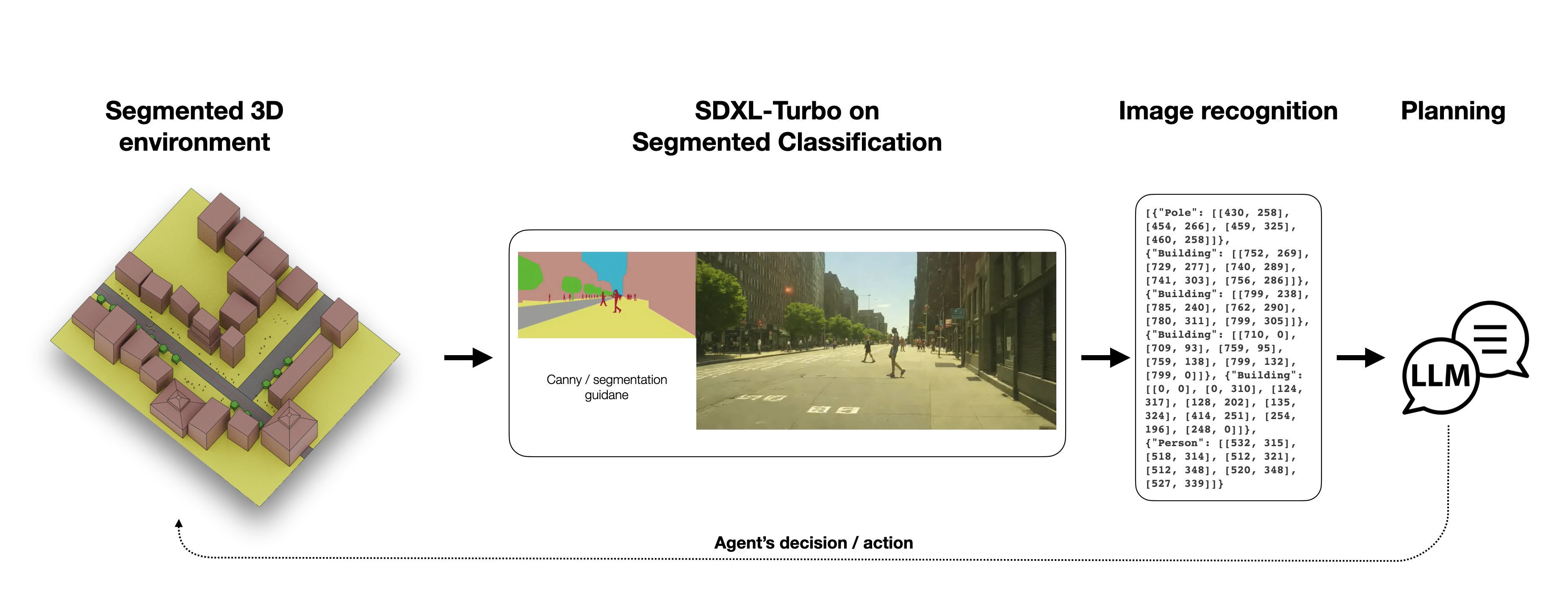}
    \caption{Pedestrian-level image generation from a 3D model. The SDXL-Turbo image generation model (middle) uses a class or canny-guided diffusion model~\citep{podell2023sdxl} to generate realistic street-level images from the 3D model (left). The generated images then analyzed by the CoT to infer the environment (middle-right), objects, and spatial layout. The agent is provided with additional inference, such as depth estimation and collision warnings, to guide its decision-making process (right).}
    \label{fig:ta-sdxl}
\end{figure}

TravelAgent is a web platform that allows spatial designers to create, simulate, and analyze Generative Agents in built environments. The platform consists of three main components: (i) a 3D environment, (ii) sensory inputs, and (iii) a Chain-of-Thought (CoT) reasoning framework, as shown in Fig. \ref{fig:ta-scheme}. TA experiments are designed as follows:
\newline
\textbf{Environment:} The agent roams a rudimentary 3D environment, which could be created using any common modeling tool with minimal details. An image generation diffusion model generates realistic street-level images from the basic 3D model to serve as input for the agent visual observation. As described in Fig.~\ref{fig:ta-sdxl}, the 3D model contains objects that are semantically segmented, so that a class-guided image generation model can reference them when generating first-person images~\citep{podell2023sdxl}.
\newline
\textbf{Agent Initialization:} Agents are initialized with a set of parameters, which can range from simple properties to complex characters and environments. At minimum, these settings include the agent's starting location, orientation, environment description, number of allowed simulation steps, and task objectives. The agent's memory can be initialized as an empty set or can contain certain priors, for example - to reference a known environment. The agent's task is defined as a set of objectives in natural language, such as reaching a destination, finding an object, or free roaming for exploration, for example - visiting a museum gallery.
\newline
\textbf{Simulation Steps:} At each step, the agent collects `sensory inputs' from the environment, such as street-level images, segmentation maps, discovery map, and memory, as described in Sec.~\ref{subsec_sensory_inputs}. The agent then processes these inputs using the CoT framework, first as observation, then as planning, and finally as action or decision. After making a decision, the agent executes it using a key-value pair response (i.e "move forward", "turn right", "finish") and a length parameter (i.e. "move 1 meter"). Table \ref{appendix:simulation_log} provides an example of the agent's interactions and decisions in several simulation steps. As shown in Fig.\ref{fig:ta-sentiment}, simulation steps might vary in length and impact. For example, a more `certain' agent might decide to move forward 50 meters in one step, when another might slightly turn left or right to look for its goal. At the end of each step, the agent updates its memory, and repeats the process until the task is completed or the simulation ends.

\subsection{Sensory Inputs}\label{subsec_sensory_inputs}
TravelAgents interact with the environment through a variety of sensory inputs, including visual cues, spatial memory, and prior knowledge. These inputs are designed to mimic pedestrian eye-level perception and decision-making processes, without the usage of navigation algorithms or top-down maps. The following sensory inputs are provided to the agent at each step of the simulation:
\newline
\textbf{Visual Perception:}\label{subsec_visual_perception}
TA uses computer vision to process visual information, such as object detection, scene segmentation, and depth estimation. We have explored a verity of image recognition models, such as YOLO, Mask2Former, and OpenAI's GPT~\citep{redmon2015you,achiam2023gpt,cheng2021masked}. Despite the advancements in image recognition, the agent's image-to-text inference is still abstract, and the textual information sometimes lacks the nuance and context needed for complex decision-making. To address these shortcomings, depth and collision information is provided via ray-casting, which the agent emits from their camera's viewpoint to nearby objects in the scene. This information is returned as class labels (`a wall', `a tree') and values (`{front : 2 meters}').
\newline
\textbf{Discovery Map:}\label{subsec_discovery_map}
In early experiments, agents tended to revisit previously explored locations unintentionally. To mitigate this, we introduced a `Discovery Map' that provides a top-down view of the environment, indicating the agent's current location, orientation, and previously-explored areas. The Discovery Map updates with each movement, reflecting the agent's new position and the areas it has already traversed. The map is incorporated into the Chain-of-Thought (CoT) as part of the agent's sensory inputs. Unlike traditional ABM, the Discovery Map is not used as an explicit pathfinding or navigation algorithm but serves only as a visual reference to the agent's spatial memory.
\newline
\textbf{Compass:}\label{subsec_compass}
In scenarios where the agent is assumed to have prior familiarity with the environment or possesses some navigational aid (such as a person holding a GPS enabled cellphone), a compass-like navigation cue provides general guidance. As shown in Fig. \ref{fig:ta-platform}, this compass is displayed at the bottom of the agent's field of view and is provided to the agent's Chain-of-Thought (CoT) framework as part of the image inference process. The compass can be omitted to simulate a completely unfamiliar environment, such as visiting a museum gallery for the first time, compelling the agent to rely solely on visual cues and spatial memory for navigation. As with the Discovery Map, the agent is not forced to navigate using the compass, and can alter its path as needed, as discussed in Section~\ref{subsec_spatial_analysis}.
\newline
\textbf{Spatial Memory:}
TA maintains a spatial memory of the environment, which is updated at each step. This memory encapsulates the agent's past experiences, including visited locations, observed objects, and navigational cues. The memory is stored in a compressed textual format, representing the agent's accumulated knowledge and experiences, and can be used to initialize new agents in future simulations, providing continuity and context. As discussed in Section~\ref{sec_results}, the agent's memory can be analyzed to understand the agent's behavior, experiences, and decision-making processes.
\newline
Despite this verity of sensory inputs, human perception of physical environments is complex and multifaceted~\citep{kim1999spatial}, and the agent's ability to interpret and respond to these inputs is limited by its current inference capabilities. With the advancement of multimodal models~\citep{reid2024gemini}, future iterations of TA should explore wider range of sensory inputs, such as audio cues and social interactions, to enhance the realism and complexity of the agent's experiences.


\section{Case Studies} \label{sec_case_studies}
To facilitate experimentation, we developed an easy to use web-based platform that enables agent creation and evaluation. We utilized this platform to conduct a series of experiments in various settings, including both interior and exterior spaces, to evaluate the effectiveness of TAs in simulating human behavior. This section presents the conducted experiments, with the results and analysis discussed in Section~\ref{sec_results}.

\begin{figure}
    \centering
    \includegraphics[width=1\textwidth]{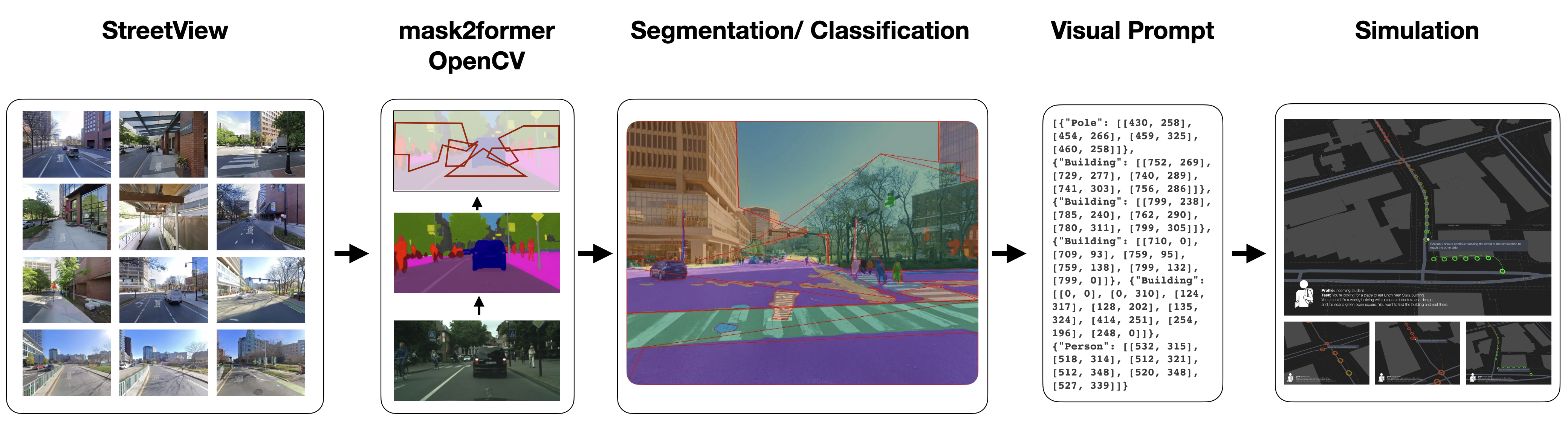}
    \caption{Visual Perception in `Lunch Break' Experiment. The agent's visual perception is guided by Google Street View (GSV) images (left), which provide a first-person view of the environment. A Mask2Former model is used to segment the image and identify objects, such as buildings, trees, and benches. An OpenCV convex hull algorithm is used to estimate the segments outlines (middle), which are then used as a textual reference for the agent's navigation and decision-making process. (right) The agent's steps are visualized on the map, with decision points marked as circles, turning red as the agent progresses.}
    \label{fig:ta-gsv}
\end{figure}

\subsection{Early Experiments}\label{subsec_early_experiments}
\textbf{Exterior - Lunch Break in Kendall Square:} An early experiment evaluated the agent's ability to navigate and make decisions in real-world environments. The scenario involved simulating a student's search for a shaded lunch spot in Kendall Square, Cambridge, MA. Without a map or predefined route, the agent relied on visual cues and memory to navigate, simulating an unfamiliarity with the area. Google Street View (GSV) images were used to represent the agent's point of view. As shown in Fig.~\ref{fig:ta-gsv}, at each step the agent is presented with a new GSV image and is asked to make decisions based on visual information, memory, and task requirements.
\newline
The use of GSV images provided realistic streetscapes but introduced certain challenges. The agent's visual perception was limited, as GSV is primarily captured from the center of the road and not in pedestrian areas. Additionally, navigation cues such as depth, distance, and collision warnings were challenging to produce, as GSV does not provide detailed object segmentation or depth information. Furthermore, GSV data cannot be easily modified to present different scenarios or environments, such as design iterations, varying seasons, times of day, or weather conditions, limiting the experiment's flexibility and scalability.
\newline
\textbf{Interior - Laboratory Exploration:} In a subsequent experiment, we simulated a navigation task within a laboratory environment. Unlike GSV, this experiment utilized a detailed 3D model of a lab, featuring specific elements such as rooms, elevators, staircases, and lab equipment. Here, the agent was tasked with meeting a colleague in a public area of the lab. In several iterations, the agent successfully navigated to the designated location but struggled with consistent decision-making and pathfinding, often revisiting the same locations or getting stuck in corners. This experiment highlighted the importance of spatial memory as part of the CoT framework, as well as the need for more detailed and context-specific sensory inputs to facilitate task-driven navigation. These findings led to the design of the sensory inputs detailed in Section~\ref{subsec_sensory_inputs}.


\subsection{Main Experiment: `Train Station'}
Building on previous studies, we designed a controlled experiment to evaluate the effectiveness of the Chain-of-Thought (CoT) framework in guiding agent behavior. Here, the primary objective was to simulate a typical daily commute, and assess how well agents could navigate urban environments to reach a designated train station. Unlike earlier experiments, this study emphasized the agent's ability to integrate task objectives with dynamic observations and memory, rather than relying on initial conditions or step-by-step navigation. Agents were provided with no maps and did not follow any shortest-path or predefined navigation routes.

\subsubsection{Experiment Design}\label{subsec_experiment_design}
\begin{figure}
    \centering
    \includegraphics[width=1\textwidth]{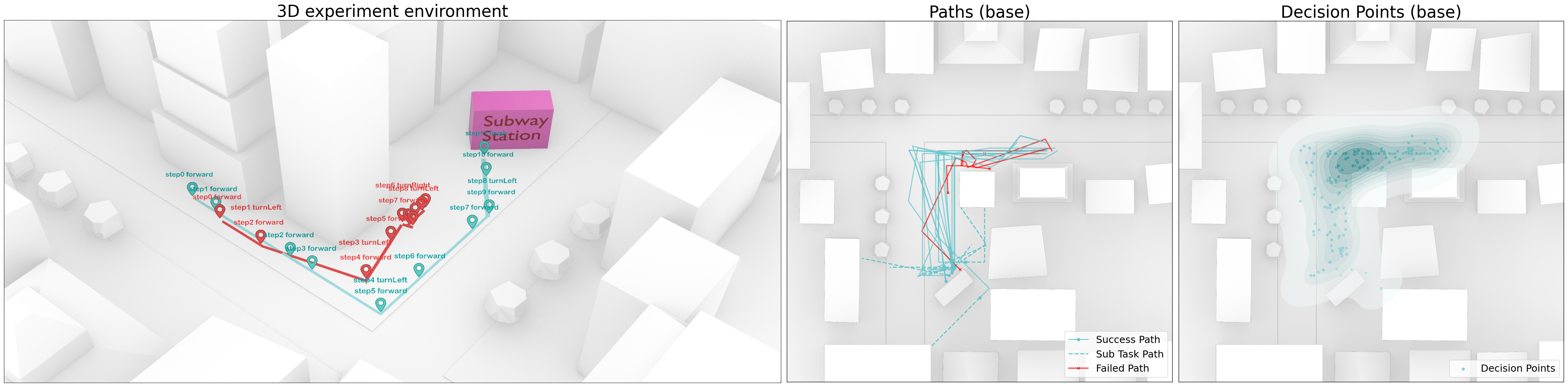}
    \includegraphics[width=1\textwidth]{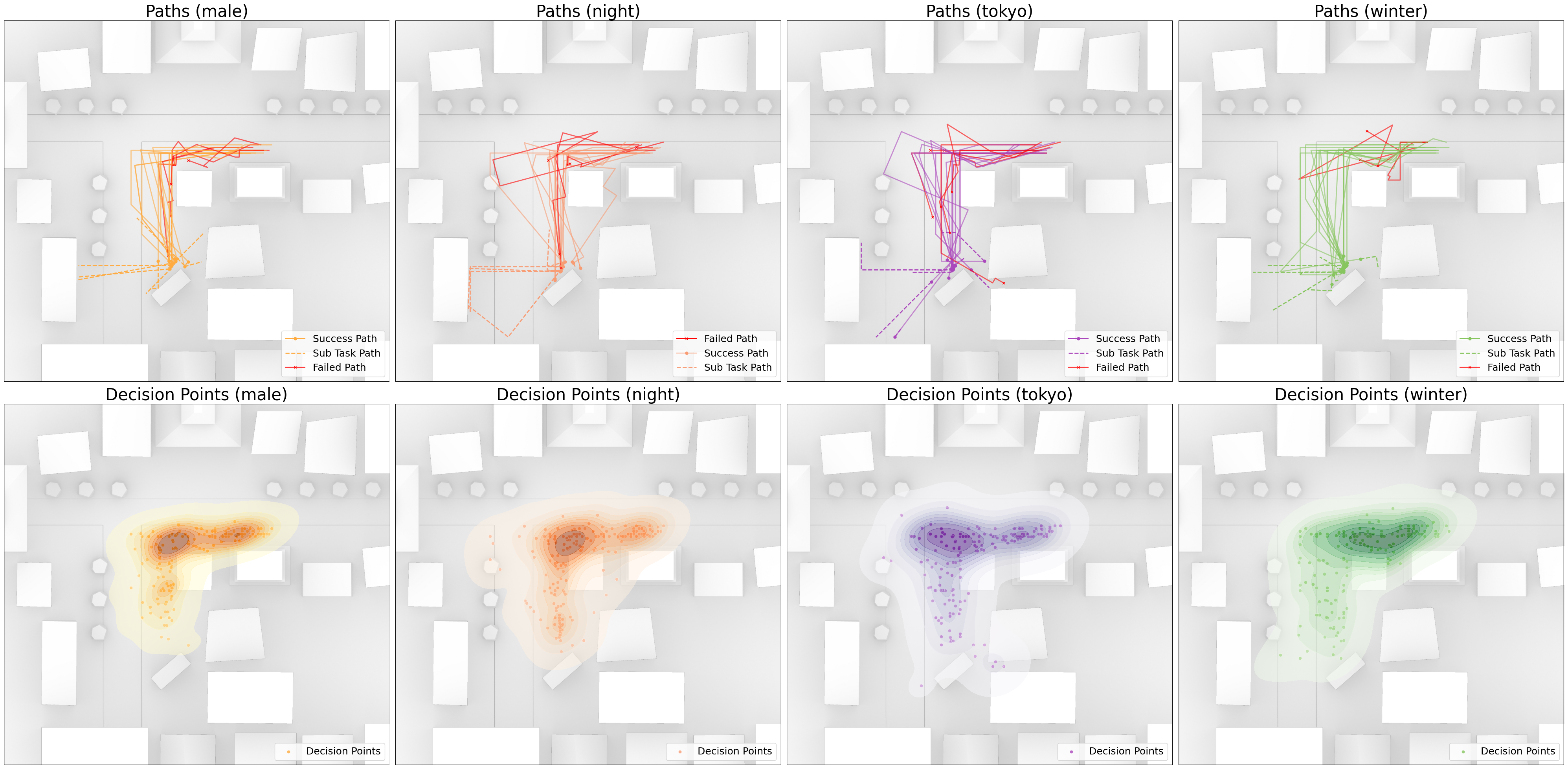}
    \caption{Spatial Analysis of `Train Station' Experiment. For each scenario, we evaluate the agent's successful and failed paths. If the agent reaches and recognizes the subway station by declaring `stop', the path is considered successful, and the agent is given a new subtask. The top left figure show both a successful and a failed path in the `Base' scenario. The top row in the bottom figure shows the paths of all agents in all other scenarios, and the bottom row shows the spatial aggregation of decision points across all scenarios. Notably, the agents in the `Night' scenario have the most failed and inconsistent paths, which is also reflected in more sparse decision point area. Conversely, the agents in the `Winter' scenario have more consistent paths, with a clear aggregation of decision points early in the simulation.}
    \label{fig:ta-paths}
\end{figure}

The experiment was structured around a consistent urban environment and the same task, but varied across different scenarios, personas, and initial conditions. The environment comprised an urban scene featuring a mix of buildings, streets, urban design elements, and pedestrians, with the task being to navigate to a nearby subway station as part of a daily commute. Each agent received a natural language prompt detailing the agent's persona, time of day, weather conditions, and task requirements. Agent's personas may be as simple as a `student' or `researcher', or more complex to include additional attributes such as `age', `gender', or certain disabilities (e.g., `wheelchair user'). The personas and scenarios were then combined to create a matrix of 100 experiments. Table~\ref{appendix:scenarios} summarizes the scenarios and table~\ref{appendix:scenarios_matrix} illustrates the associations between season, location, time, persona, and scene. Upon successful completion of its task, the agent was informed that the train service was unavailable, and was subsequently assigned an additional task from this set of sub-tasks:
\begin{itemize}\label{list_subtasks}
    \item `If the train is unavailable, find an alternative way to get to work.'
    \item `Buy coffee before work if there's time.'
    \item `Interact with a friend across the street if encountered.'
\end{itemize}
These sub-tasks were designed to evaluate the agent's adaptability to unforeseen circumstances and new information. The agent's action in these sub-tasks is shown as dotted lines in Fig.~\ref{fig:ta-paths}.

\subsubsection{Agent Initation}
In this experiment, all agents are aquatinted with their environment. To further reinforce their familiarity, the initiation prompt includes: \textit{`You reside on this street. It is morning, and you are on your way to work. Your objective is to reach the nearby subway station. To reach the station, proceed down the street, then turn left, and it will be on your left. Look for a large `Subway' sign.'} This prompt is designed to provide the agent with a general sense of direction and task objectives, without explicitly guiding them through a predefined path or route. In unfamiliar environments, the agent is provided with a more general prompt, such as \textit{`You are visiting this gallery for the first time. You are here to view the new exhibit. Find the exhibit hall and explore the rest of the museum.'}

\section{Results} \label{sec_results}
In this section we present the results of the case studies, focusing on the `Daily Commute' experiment. We discuss the agent's behavior, decision-making process, and the opportunities and challenges of using TravelAgent.

\subsection{Results: Early Experiments}
In the initial experiments (see~\ref{subsec_early_experiments}), the agent's decisions were primarily influenced by immediate visual stimuli and physical elements recognized by the computer vision model. In all tests, the agent's path was largely determined by early choices, such as its initial direction or its decision at a fork, leading to a deterministic behavior pattern that could not be attributed to any cognitive process. This suggests a strong influence of initial conditions, with limited adaptation to new information, and minimal reliance on memory. To address this, subsequent simulations incorporated and enhanced the CoT, allowing the agent to re-evaluate choices based on updated sensory inputs and environmental cues.

\subsection{Results: `Subway Station'}
In our primary experiment we conducted 100 simulations totaling 1,898 steps. Approximately 76\% of the agents successfully completed their tasks within the stipulated step count. The remaining 24\% failed to achieve their objectives due to factors such as obstacle encounters or disorientation caused by continuous turning maneuvers. In this section we present a comprehensive evaluation of the agents' behaviors and decision-making processes using spatial analysis, thematic and topical modeling, and sentiment analysis of the agents' cognitive outputs and observations. Appendix~\ref{appendix:simulation_log} provides a log of the agent's interactions and decisions in one iteration of the 'Subway Station' experiment.
\newline
It is important to emphasize that reaching the goal (i.e., finding the subway station) was not the main objective of these experiments. Instead, TA is designed to inspect the agent's observation, planning, and decisions, so that the legibility and coherence of the environmental design could be evaluated based on these findings. In that respect, `failed' experiments may tell more about the spatial design characteristics than successful ones.

\subsection{Spatial Analysis}\label{subsec_spatial_analysis}
An evaluation of paths across different scenarios in the `Subway Station' experiment revealed distinct spatial patterns, as shown in Fig.~\ref{fig:ta-paths} (bottom). Agents in the `Night' scenario exhibited the highest frequency of navigation failures and the most inconsistent path trajectories. This is reflected in a more dispersed aggregation of decision points, suggesting difficulties in spatial orientation and decision-making under reduced visibility conditions. Conversely, agents in the `Winter' scenario demonstrated more consistent and successful navigation paths, with decision points clustered tightly and earlier in the simulation. Correlating with the top-terms analysis (see Sec.~\ref{subsec_term_frequency}), the agents in the 'Winter' scenario focused on the presence of snow, ice, and cold weather, which may have influenced their tight navigation strategies and goal-oriented decision-making process.
\newline
As described in Sec.~\ref{subsec_experiment_design}, upon reaching the subway station, the agent was assigned an additional sub-task. As shown in~\ref{appendix:simulation_log}, the agent's behavior during these sub-tasks is more exploratory and less deterministic, as the agent is not provided with a clear objective or navigational cues to follow. Still, the agent is able to adapt to new information and make decisions based on its prior observations and memory. For instance, the agent concludes that a `coffee shop' is likely to be located near the subway station, where it located a plaza. This demonstrates the agent's ability to infer and reason about its environment as well as rely on its memory, even in the absence of explicit instructions.

\begin{figure}
    \centering
    \includegraphics[width=.75\textwidth]{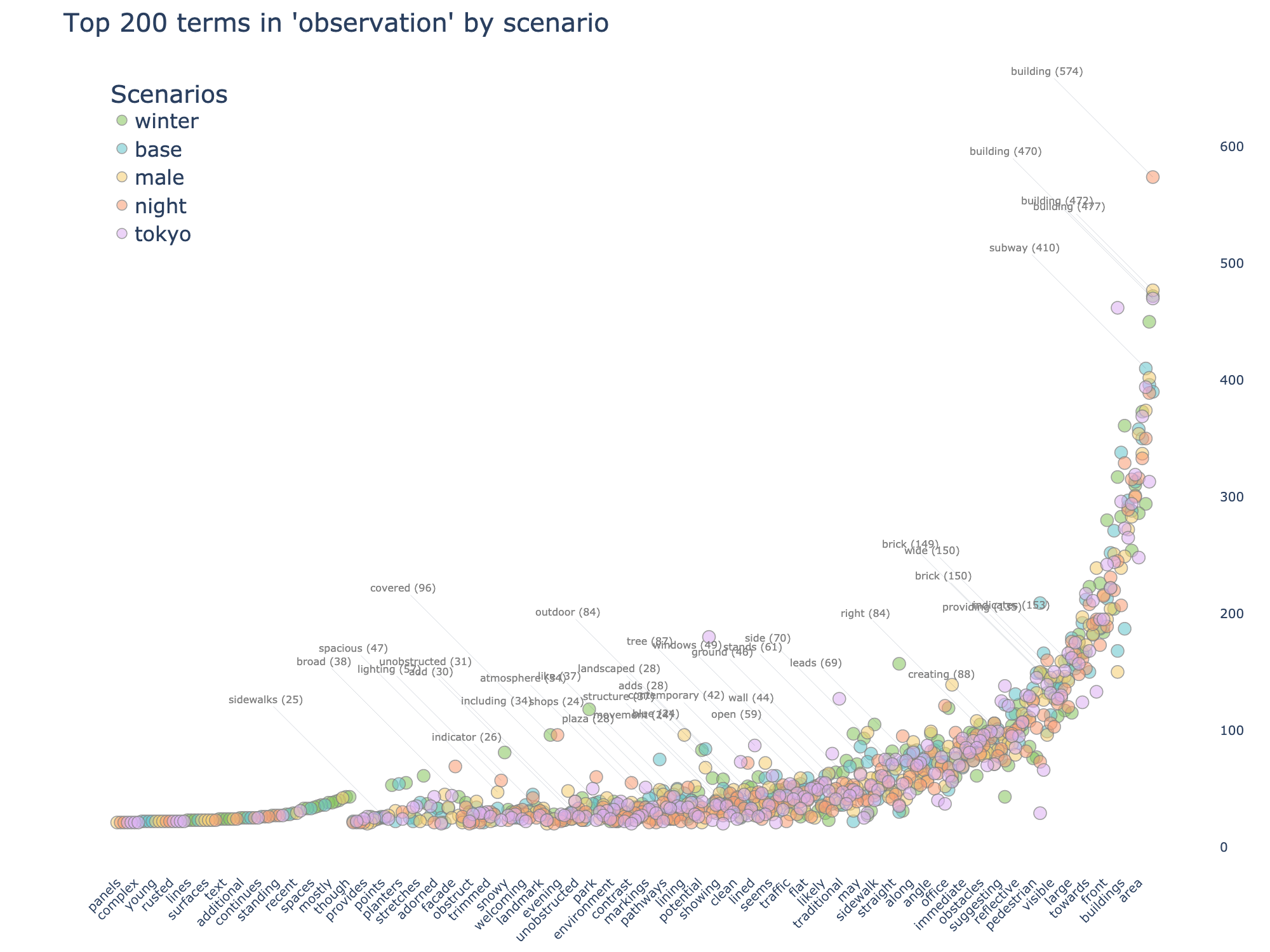}
    \caption{Analysis of the 200 most frequent words in the agents' planning streams, clustered by scenario. The dominant terms correspond to the agents' tasks, goals, actions, and interactions with their immediate surroundings. Across all scenarios, the most prevalent terms are related to the agents' target and goal, such as `station', `target', and `move'. The agents also reflect on the shape, form, and materiality of their environment, with terms like `bricks', `glass', `flat', and `building' appearing frequently.}
    \label{fig:ta-top-words}
\end{figure}

\subsection{Term Frequency Analysis}\label{subsec_term_frequency}
To further understand the agent's cognitive process, we conducted a thematic analysis of its outputs and environmental observations. We extracted the agent's internal representations from the simulation logs and applied various natural language processing (NLP) techniques, including tokenization, lemmatization, and n-gram analysis. Subsequently, we utilized term frequency-inverse document frequency (TF-IDF) vectorization to identify the most prevalent features in the agent's outputs~\citep{kantardzic2011data}. These features were then clustered to show the key topics that emerged from the agent's decision-making process.
\newline
We evaluated 1,898 unique steps in the simulation logs. From these, we extracted the following word counts: Planning - 91,340, Memory: 156,663, Observations: 219,742. The results visualized in Fig.~\ref{fig:ta-top-words}, illustrate the most prevalent terms in the agent's `plan' stream, clustered by scenario. Top features are mostly related to the agent's immediate surrounding environment and its task, such as `station', `target', `action', `move', and `street'. In all scenarios, the agent reflects on the shape, form, and materiality of its environment (`bricks', `glass', `flat', `building'), as well as the presence of obstacles and navigational cues (`obstacle').

\begin{figure}[ht]
    \centering
    \includegraphics[width=1\textwidth]{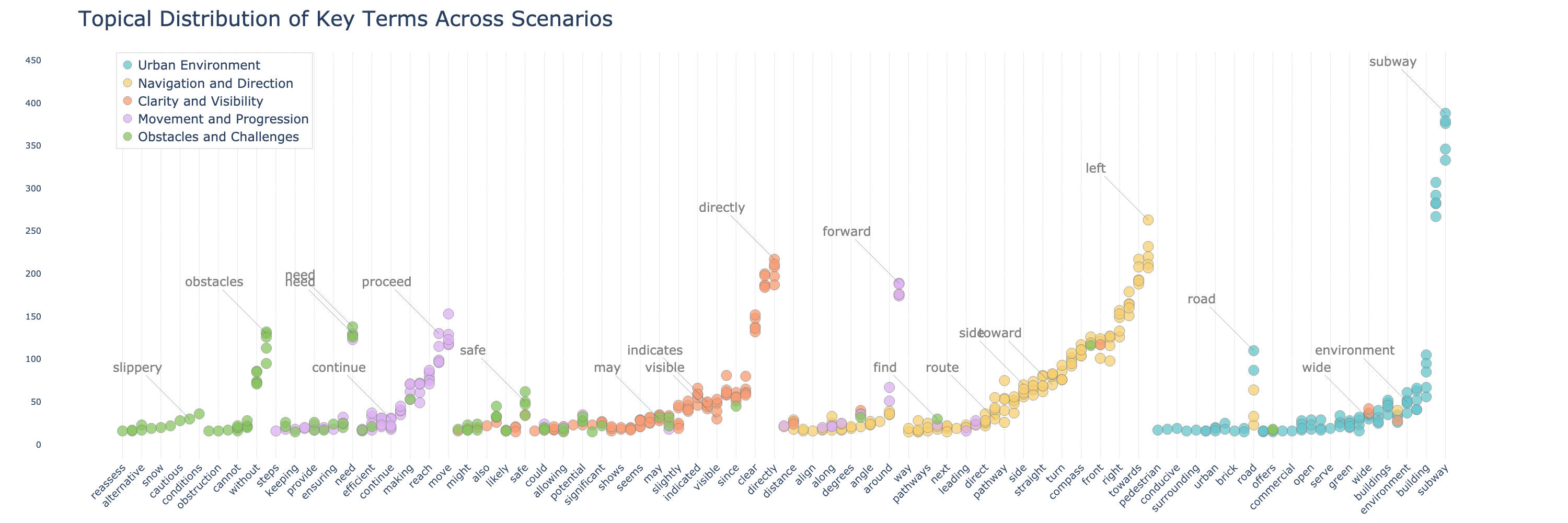}
    \caption{Topical modeling of agent observations and plans across scenarios. Here, key terms in the agent's observations and plans (as discussed in~\ref{fig:ta-top-words}) are clustered into five major topics: navigation, visibility, movement, obstacles, and urban environment. An NLP method is used to extract the main topics from the entire corpus of the agent's plans and observations. Each top term is then assigned a topic based on its semantic content, and the distribution of topics is visualized across all scenarios.}
    \label{fig:ta-topical}
\end{figure}

\subsection{Topical Modeling}
To examine the cognitive processes underlying the agent's behavior, we conducted topic modeling on the observation and planning streams. Key topics were extracted using LLM clustering and semantic analysis~\citep{petukhova2024text}, categorizing terms into navigation, visibility, movement, obstacles, and urban environment. Latent Dirichlet allocation (LDA) was applied to the agent's logs to identify semantic themes~\citep{blei2003latent}. As depicted in Fig.~\ref{fig:ta-topical}, the analysis revealed a predominance of terms related to urban features and the agent's navigation, with less emphasis on movement, obstacles, and visibility. This model indicated a strong focus on direct path-following and target-seeking behaviors, with limited exploration or route optimization. The Natural Language Toolkit (NLTK) and GPT-4 were utilized for textual analysis and initial semantic categorization~\citep{achiam2023gpt,hutto2014vader}.
\newline
As shown in Fig.~\ref{fig:ta-paths}, the 3D environment for this experiment was designed to include potential shortcuts and detours, to allow for variability in the agent's decisions. Despite these features, the agent consistently followed the main road, indicating a deterministic decision-making process. The agent's behavior may be influenced by its initial conditions, prompting reliance on prior knowledge and the navigational cues (discovery map and compass), rather than adapting to new opportunities. As discussed in Sec.~\ref{sec_discussion}, further research is required to explore the agent's adaptability to dynamic and open-ended environments, including changing weather conditions, pedestrian interactions, or unexpected obstacles.

\begin{figure}
    \centering
    \includegraphics[width=1\textwidth]{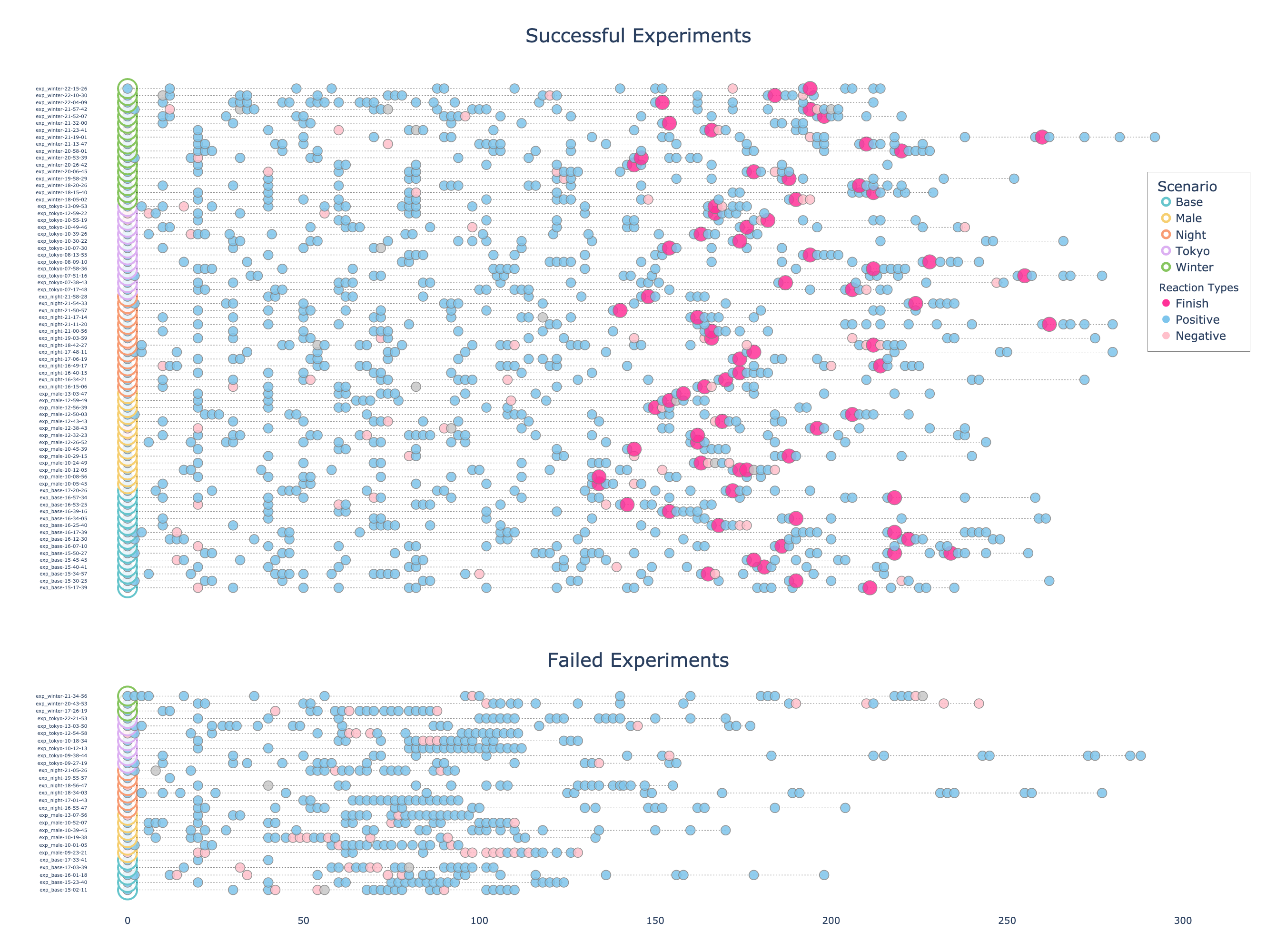}
    \caption{Sentiment analysis of the agent's `thoughts' and `observations' streams. Each experiment is displayed as a path with the agent's decision points. The distance between each decision point refers to the distance traveled in `Forward' actions. Sentiments are classified into three categories: positive (blue), neutral (gray), and negative (light pink) using NLTK’s VADER~\citep{hutto2014vader}. In successful experiments, the red dots represent the "Finish" steps, where the agent is given an additional randomized subtask. Notably, in failed paths, there are clear clusters of `search' actions, in which the agent aims to reorient itself, usually leading to a negative sentiment. In successful paths, the agent's sentiment is more positive, and search steps are less frequent. In sub-tasks, mixed sentiments appear, alongside some `search' clusters.}
    \label{fig:ta-sentiment}
\end{figure}

\subsection{Sentiment Analysis}
To evaluate the agent's `emotional' state in both successful and failed paths, we conducted sentiment analysis using the Natural Language Toolkit's (NLTK) VADER sentiment analyzer~\citep{hutto2014vader}. The agent's `thought' and `observation' streams were classified into three categories: positive, neutral, and negative. The results, visualized in Fig.~\ref{fig:ta-sentiment}, indicate a distinction between successful and failed paths. In both cases, the agent's sentiment is mostly positive, and occasional negative sentiment can be observed. However, failed paths exhibit clusters of both negative sentiments and iterative actions (`search' steps), indicating the agent's `frustration' or `confusion' when failing to find a path. In contrast, successful paths show a more positive sentiment, with fewer search steps and a clearer path to the target.
\newline
As described in \ref{list_subtasks}, the agent is assigned an additional task upon reaching its goal. During these additional tasks, the agent's sentiment tends to be more negative and `search' oriented, reflecting a sense of confusion. As shown in Fig.~\ref{fig:ta-paths}, the agent's path during these sub-tasks is generally more exploratory, suggesting a lack of clear objectives or guidance. In the subtask \textit{`Interact with a friend across the street'}, the agent's sentiment appears more positive, and its actions are more deliberate. This might indicate that the agent can be more engaged and motivated when presented with a less ambiguous task, even in the absence of explicit instructions or navigational cues.
\newline
This sentiment analysis is conducted directly on the agent's `thought' and `observation'. For that reason, it is likely that emotional states are not fully captured, and the agent's sentiment is more reflective of its immediate environment and task. Future research should explore additional steps in the agent's CoT framework to incorporate more nuanced emotional states. For example, the agent could be requested to reflect on its `emotions' or `motivation' to perform certain actions, providing a more comprehensive understanding of the agent's cognitive processes.

\section{Discussion} \label{sec_discussion}
This paper presents TravelAgent (TA), a novel platform for simulating human behavior in urban environments using generative agents and a Chain-of-Thought (CoT) framework. TA models decision-making, experiences, and interactions of agents in virtual settings, providing urban designers, planners, and architects with a new way for evaluating spatial designs. Experiments presented here demonstrate TA's effectiveness in simulating behavior and assessing agents' interactions. This section discusses TA's potential implications for spatial design, along with the opportunities and challenges of using generative agents for behavior simulation.

\subsection{Implications for Urban Design and Architecture}
TA has several potential implications for urban design and architecture, particularly in the areas of wayfinding, environmental legibility, and user experience.
\newline
\textit{Wayfinding and Navigation:} TA's ability to simulate human-like navigation offers insights into urban interactions. By analyzing agent paths and decision points, designers can identify wayfinding challenges and optimize layouts for better navigability. For instance, the clustering of decision points in the `Winter' scenario highlights the importance of clear sightlines and distinctive landmarks in aiding navigation. Inversely, the sparse decision points in the `Night' scenario can suggest the need for improved signage, lighting, and visibility. Integrating these elements can enhance wayfinding and reduce cognitive load on pedestrians in urban environments.
\newline
\textit{Environmental Legibility:} TA simulations underscore the significance of visual cues and spatial memory in navigation. By analyzing agents' observations and plans, designers can evaluate the legibility of various design options. Clear landmarks and visual cues, as indicated by the agents' cognitive outputs, enhance the clarity and coherence of urban spaces. Materials, opacities, and textures can also play a key role in navigation tasks and spatial orientation, as discussed in Section~\ref{subsec_term_frequency}.
\newline
\textit{User Experience and Safety:} Sentiment analysis can identify areas where agents experience frustration or confusion, and help address safety concerns and enhance user experience. For instance, negative sentiments and search steps in failed paths indicate confusing areas; Lack of clear signage, obstructed views, or ambiguous pathways may contribute to user discomfort and disorientation. Simulating diverse agents across demographics and personas can help designers assess inclusivity and accessibility, leading to more user-friendly and safe urban environments.

\subsection{Limitations and Future Work}
While TA demonstrates potential in simulating generative agent behavior within urban environments, several limitations currently constrain its effectiveness. This section delineates these key limitations and proposes future research directions to enhance the platform's capabilities.
\newline
\textit{Validation and Real-World Integration:} Despite TA potential in simulating human-like behavior, it is crucial to distinguish between the agent's seemingly natural actions and actual human behavior. The agent's decisions are heavily influenced by its training data and initial simulation conditions. Current research highlights the importance of addressing issues such as accuracy, bias, hallucinations, and generalization in LLMs, which underpin TA's reasoning and decision-making processes~\citep{ferrara2023should,kotek2023gender}. In urban environments, TA agents may exhibit further biases or unrealistic behaviors that could affect their perception and decision-making~\citep{manvi2024large}. Extensive validation is necessary to assess TA's accuracy and reliability. Future research should involve comparing TA simulations with human-subject studies~\cite{gath2024role}, real-world observations, and traditional ABMs, as well as conducting user studies to evaluate the platform's effectiveness and applicability.
\newline
\textit{Agent Diversity and Personalization:} TA agents may not fully capture the diversity of individuals and groups. Future work should develop more sophisticated agent profiles reflecting varying demographics, such as age, mobility, cultural background, and preferences. Simulating agents representing elderly users, individuals with disabilities, or children could enable designers to evaluate the accessibility and inclusivity of their spatial design.
\newline
\textit{Environmental Complexity and Dynamics:} The current TA implementation is limited to simple environments with basic goals. Future work should incorporate more complex and dynamic environments, such as diverse building types, varying weather conditions, and fluctuating pedestrian and vehicular traffic. The agent's action itself is limited to simple movements, such as `move forward' or `turn left', without considering more nuanced behaviors, such as interacting with objects or people. Future research should explore more sophisticated agent interactions, including multi-step tasks, object manipulation, as well as 3D movement (e.g., climbing stairs, riding elevators). Moreover, enriching the platform with multi-agent interactions would enable the study of social dynamics and collective behaviors, such as crowd movements and pedestrian flows. Incorporating agent-to-agent communication and group dynamics can better mirror human interactions in urban settings, and its effect on decision-making.
\newline
\textit{Computational Efficiency:} TA currently requires significant computational resources for generating sensory inputs, processing behaviors, and performing CoT analysis. Future efforts should focus on optimizing performance to reduce simulation time and resource demands, potentially by parallelizing processes and offloading some decision-making to traditional ABM logic. Additionally, generative streetscapes images still lack consistency across steps. Future work should focus on improving the image generation model to provide more realistic and coherent visual inputs to the agent, potentially by incorporating generation techniques that focus on temporal consistency.
\newline
\textit{Applications in Urban Safety and Policy Planning:} Future TA research should explore and assess design options for improving urban safety and security. By simulating emergency scenarios, such as evacuations or disaster response, TA can help identify vulnerabilities and optimize urban layouts for safety. Mixing TA with traditional ABMs can provide an enhanced view of urban dynamics and inform policy planning and decision-making. In addition, simulating the impact of policy interventions, such as changes in zoning regulations, transportation infrastructure, and public space design, can help policymakers anticipate outcomes and make informed decisions.

\subsection{Conclusion}
TravelAgent (TA) is a novel platform designed to simulate human behavior in urban environments by integrating pedestrian-level generative agents and a Chain-of-Thought (CoT) framework. TA enables designers, planners, and architects to evaluate spatial configurations, asses user interactions, and enhance the performance and functionality of their designed spaces. Experimental results demonstrate TA's potential to provide valuable insights into wayfinding, environmental legibility, and user experience. Despite its promise, TA's current implementation requires further validation, increased environmental complexity, improved computational efficiency, and greater agent diversity. Addressing these challenges will enable future iterations of TA to become an integral part of the designer's toolbox, allowing designers to evaluate and optimize their designs based on human-centric criteria and agent-driven feedback.

\newpage

\bibliography{references}
\bibliographystyle{unsrtnat}

\newpage

\section*{Appendix}
\subsection*{Appendix A: Scenario Descriptions}

\begin{itemize} \label{appendix:scenarios}
    \item \textit{Scenario 1 - `Base':} A bright summer morning in Kendall Square, Cambridge, MA. Modern glass buildings and bustling streets with pedestrians, cyclists, and outdoor cafés creating a lively and vibrant atmosphere.
    \item \textit{Scenario 2 - `Winter':} A snowy winter morning in Kendall Square, Cambridge, MA. Heavy snow blankets the streets and modern buildings, with bundled-up pedestrians, and a quiet stillness filling the air.
    \item \textit{Scenario 3 -`Tokyo':} A bright summer morning in Tokyo, Japan. Mix of traditional wooden buildings and modern structures. Bustling streets filled with pedestrians, cyclists, and outdoor tea houses creating a vibrant atmosphere.
    \item \textit{Scenario 4 - `Night':} A quite summer nighttime street scape in Kendall Square, Cambridge, MA. Streets are softly illuminated by warm streetlights and the gentle glow of modern office buildings with large glass facades.
    \item \textit{Scenario 5 - `Persona':} The agent's persona was varied between a 30-year-old female and a male around the same age. This was meant to examine wether the agent's gender influenced its decision-making process.
\end{itemize}

\subsection*{Appendix B: Scenario Matrix}

\begin{table}[!ht]
    \centering
    \begin{tabular}{@{}llcll@{}}
        \toprule
        \textbf{Season} & \textbf{Location}      & \textbf{Time} & \textbf{Persona}        & \textbf{Scene} \\
        \midrule
        Summer          & Kendall Square, Boston & Morning       & 30-year-old female/male & Scene 1        \\
        Winter          & Kendall Square, Boston & Morning       & 30-year-old female/male & Scene 2        \\
        Summer          & Tokyo                  & Morning       & 30-year-old female/male & Scene 3        \\
        Summer          & Kendall Square, Boston & Night         & 30-year-old female/male & Scene 4        \\
        \bottomrule
    \end{tabular}

    \caption{Associations between Season, Location, Time, Persona, and Scene \label{appendix:scenarios_matrix}}
\end{table}

\subsection*{Appendix C: Simulation Log}\label{appendix:simulation_log}

This section presents partial row output from experiment `exp\_night-17-48-11' in the `Train Station' study. The following table shows selected steps from the simulation log, which include both segmented and generated street views, task, observation, plan, memory, warning, and action. The steps displayed are from the beginning of the simulation [0,1], later when the agent reaches the subway station [11], and finally when it is given an additional task [12, 15]. The additional task is to `Find a coffee shop before work if there's time'.
Although a location of the coffee shop is not provided, the agent is visually evaluating its surroundings, and assesses that a coffee shop is likely to be in the vicinity of the subway station, where a plaza is recognized:
\newline
\textit{``To efficiently reach the coffee shop, I should proceed forward towards the plaza, as it is likely to have coffee offerings among the commercial enterprises. The right side is blocked, making leftward or forward movement essential. Moving towards the center will allow me to explore the plaza without retracing steps unnecessarily''}.
\newline
This demonstrates the agent's ability to infer and plan based on its common knowledge and observations, despite the lack of explicit instructions. The rest of the simulation log is available below.

\input{appendix/logs_table.tex}

\end{document}

%% file: appendix/logs_table.tex
\begin{longtable}{|p{1\textwidth}|}
    \caption{Steps and Corresponding Messages}                                                                                                                                                                                                                                                                                                                                                                                                                                                                                                                                                                                                                                                                                                                                                                                                                                                                                                                                                                                                                                                                                                                                                    \\
    \hline
    \textbf{Step and Messages}                                                                                                                                                                                                                                                                                                                                                                                                                                                                                                                                                                                                                                                                                                                                                                                                                                                                                                                                                                                                                                                                                                                                                                    \\
    \hline
    \endfirsthead

    \multicolumn{1}{c}{\tablename\ \thetable\ -- Continued from previous page}                                                                                                                                                                                                                                                                                                                                                                                                                                                                                                                                                                                                                                                                                                                                                                                                                                                                                                                                                                                                                                                                                                                    \\
    \hline
    \textbf{Step and Messages}                                                                                                                                                                                                                                                                                                                                                                                                                                                                                                                                                                                                                                                                                                                                                                                                                                                                                                                                                                                                                                                                                                                                                                    \\
    \hline
    \endhead

    \hline \multicolumn{1}{|r|}{Continued on next page}                                                                                                                                                                                                                                                                                                                                                                                                                                                                                                                                                                                                                                                                                                                                                                                                                                                                                                                                                                                                                                                                                                                                           \\
    \endfoot

    \hline
    \endlastfoot
    \begin{minipage}[t]{\linewidth}\centering
        \par\vspace{0.1em}
        \includegraphics[width=0.5\textwidth]{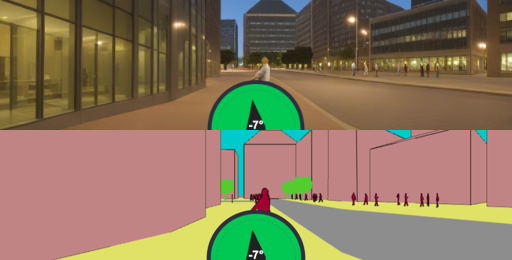}
    \end{minipage}\par\vspace{0.5em}

    \begin{minipage}[t]{\linewidth}
        \scriptsize
        \textbf{Log for step 0}\par\vspace{0.5em}
        \textbf{Task:} Navigating through the street to reach the 'Subway Station' sign.\par\vspace{0.5em}
        \textbf{Plan:} {"thought":"Given that the subway station sign is located off to the right at a slight angle, I need to adjust my position to navigate toward it while avoiding any obstacles in my path. The clear road ahead allows me to move efficiently in that direction."}\par\vspace{0.5em}
        \textbf{Memory:} {"previous\_observations":"While navigating the simulated environment, I observed a series of distinct landmarks including a large, ancient tree at the center of a clearing, a winding river to the east, and a mountain range visible to the northwest. The landscape appears largely untouched since my last interaction, providing clear paths for navigation and possible routes to explore. I noted several trails branching off towards the river and mountains, which could facilitate further exploration.","previous\_actions":"I initiated my exploration by moving forward through the clearing towards the ancient tree. After reaching it, I assessed my surroundings to determine the best route for future navigation and potential task completion."}\par\vspace{0.5em}
        \textbf{Warning:}   There is a/an building in your left in 4.61 m. \par\vspace{0.5em}
        \textbf{Observation:} {"observation":{"left":"To the left, there's a tall glass building reflecting the evening light, which looks like an office space. Beside it, a well-maintained sidewalk has some potted plants lining the edge, adding greenery to the urban environment. Further down, I see a low barrier and a few park benches, indicating a small public resting area.","front":"In the central view, the road is wide with smooth pavement, well-marked lanes, and a crosswalk ahead. The area is brightly lit with street lamps, creating a welcoming ambiance. To the right, I can see a large building with an interesting architectural design that looks to house mixed-use commercial spaces. The visibility is good with no immediate obstructions blocking my path.","right":"To the right, there are several buildings of varying heights, with the tallest being a glass and steel skyscraper. Below the walkway, I can see a parking lot with clear markings and some landscaping, including several small trees dotting the area.","compass":"The compass is showing a red light, indicating the target is to my right at an angle of 7 degrees."}}\par\vspace{0.5em}
        \textbf{Action:} {"action":"turnRight","value":15}\par\vspace{0.1em}
    \end{minipage} \\
    \hline
    \begin{minipage}[t]{\linewidth}\centering
        \par\vspace{0.1em}
        \includegraphics[width=0.5\textwidth]{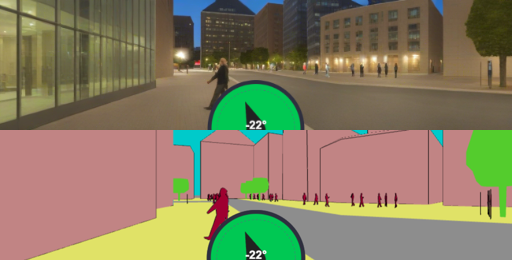}
    \end{minipage}\par\vspace{0.5em}

    \begin{minipage}[t]{\linewidth}
        \scriptsize
        \textbf{Log for step 1}\par\vspace{0.5em}
        \textbf{Task:} Navigate towards the 'Subway Station' sign, using the details in the environment and the compass for guidance.\par\vspace{0.5em}
        \textbf{Plan:} {"thought":"The subway station sign is located to the right at a -22 degree angle from my current position. Given the urban layout and my previous observations, my best option is to adjust my direction towards the right and move forward cautiously while keeping an eye on surroundings for the Subway Station sign."}\par\vspace{0.5em}
        \textbf{Memory:} {"previous\_observations":"To the left, there's a tall glass office building reflecting the evening light, with a well-maintained sidewalk adorned with potted plants. A small public resting area with park benches can be seen beyond a low barrier. In front, the wide road has smooth pavement and well-marked lanes, with a crosswalk visible ahead. The area is well-lit by street lamps. To the right, an architecturally interesting large building contains mixed-use commercial spaces, and beyond that, a variety of buildings in height, including a tall glass and steel skyscraper and a parking lot with landscaping.","previous\_actions":"[step1: turnRight(15)]"}\par\vspace{0.5em}
        \textbf{Warning:}   There is a/an building in your left in 4.77 m. \par\vspace{0.5em}
        \textbf{Observation:} {"observation":{"left":"To the left, there's a tall glass office building reflecting the evening light, with a well-maintained sidewalk adorned with potted plants. A small public resting area with park benches can be seen beyond a low barrier.","front":"The center area features a wide road with smooth pavement and well-marked lanes. A crosswalk is visible ahead, indicating a safe path to cross.","right":"To the right is an architecturally interesting large building with mixed-use commercial spaces. Beyond that, various buildings of different heights are present, including a tall glass and steel skyscraper.","compass":"The compass indicates a red direction, showing the target 'Subway Station' is to the right at an angle of -22 degrees."}}\par\vspace{0.5em}
        \textbf{Action:} {"action":"turnRight","value":22}\par\vspace{0.1em}
    \end{minipage}                                                                                                                                                                                                                                                                                                                                                                            \\
    \hline
    \begin{minipage}[t]{\linewidth}\centering
        \par\vspace{0.1em}
        \includegraphics[width=0.5\textwidth]{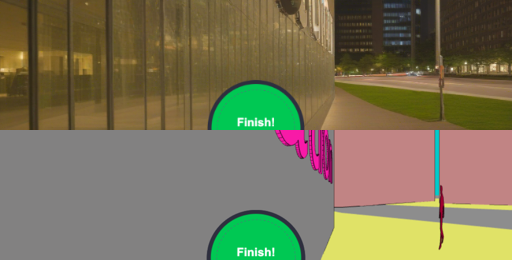}
    \end{minipage}\par\vspace{0.5em}

    \begin{minipage}[t]{\linewidth}
        \scriptsize
        \textbf{Log for step 11}\par\vspace{0.5em}
        \textbf{Task:} Reached close to the sign of the subway station. The task is now complete as the subway station is within close proximity.\par\vspace{0.5em}
        \textbf{Plan:} {"thought":"The subway station is clearly visible ahead and unobstructed. The task is completed as I'm very close to the desired target, which is the subway station sign."}\par\vspace{0.5em}
        \textbf{Memory:} {"previous\_observations":"In my recent observations, I noted continued visibility of tall glass office buildings with modern designs. The left side features a public resting area with benches and landscaped greenery, while the area straight ahead reveals a well-marked, wide road leading towards the subway station, which is free of obstacles. The right side showcases bustling mixed-use commercial buildings, enhancing the urban atmosphere. Overall, the environment remains consistent with well-maintained infrastructure and vibrant urban features.","previous\_actions":"[step1: forward(20), step2: forward(40), step3: forward(40)]"}\par\vspace{0.5em}
        \textbf{Warning:} Warning! your can't move forward.There is a/an subway station in your forward in 1.41 m.Congratulations! You have finished the task! Reply 'finish' and a value of '1' to stop the experiment.  Warning! your can't move left.There is a/an subway station in your left in 0.82 m.Congratulations! You have finished the task! Reply 'finish' and a value of '1' to stop the experiment. \par\vspace{0.5em}
        \textbf{Observation:} {"observation":{"left":"On the left, there are tall glass office buildings reflecting city lights, alongside a public resting area featuring benches and some lush landscaping, providing a pleasant break from the urban environment.","front":"In the center, the path is clear; a well-marked wide road leads directly to the subway station with clear signage indicating its presence. The road is clean and free of obstacles, making navigation straightforward.","right":"The right side is filled with bustling mixed-use commercial buildings, showcasing shops and cafes, contributing to a lively urban atmosphere as well as distractions to the navigation.","compass":"The compass indicates green, meaning the subway station is directly in front of me."}}\par\vspace{0.5em}
        \textbf{Action:} {"action":"finish","value":1}\par\vspace{0.1em}
    \end{minipage}                                                                                                                                                                                                                                                                                                                                                                           \\
    \hline
    \begin{minipage}[t]{\linewidth}\centering
        \par\vspace{0.1em}
        \includegraphics[width=0.5\textwidth]{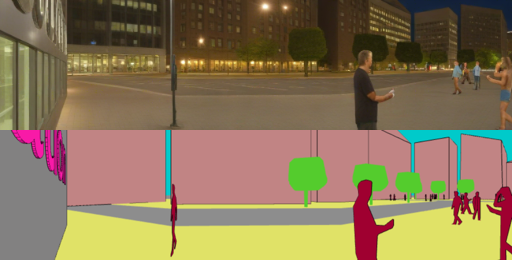}
    \end{minipage}\par\vspace{0.5em}

    \begin{minipage}[t]{\linewidth}
        \scriptsize
        \textbf{Log for step 12}\par\vspace{0.5em}
        \textbf{Task:} Analyzed environment and identified optimal navigation route for coffee purchase. Planning next steps to reach the target location efficiently and safely.\par\vspace{0.5em}
        \textbf{Plan:} {"thought":"To reach the coffee shop, I need to move forward on the clear path in front of me. The environment is safe, allowing for direct navigation without obstacles. Given that my left side is blocked by a subway station, my best option is to proceed forward toward the coffee shop indicated on my map."}\par\vspace{0.5em}
        \textbf{Memory:} {"previous\_observations":"There are several tall glass office buildings on the left, contributing to a modern urban landscape with areas for public seating. The road ahead remains well-marked and leads directly to a subway station, facilitating easy navigation. On the right, various lively mixed-use commercial buildings add to the vibrant city atmosphere, creating potential distractions while navigating.","previous\_actions":"[step1: forward(40), step2: finish(1), step3: turnRight(90)]"}\par\vspace{0.5em}
        \textbf{Warning:}  Warning! your can't move backward.There is a/an subway station in your backward in 0.82 m. Warning! your can't move left.There is a/an subway station in your left in 1.41 m. \par\vspace{0.5em}
        \textbf{Observation:} {"observation":{"left":"The left side features several tall glass office buildings, creating a modern aesthetic. There are areas designed for public seating among landscaped greenery, promoting a welcoming atmosphere.","front":"The central view shows a well-marked road leading directly ahead, unobstructed and clear for movement. The road appears in good condition with appropriate traffic markings, indicating safe navigation toward the target.","right":"To the right are mixed-use commercial buildings with storefronts, adding vibrancy to the area. These buildings are of varying heights and styles, enhancing the urban feel. The lighting conditions suggest it is early evening, contributing positively to the overall ambience.","compass":"there is no compass information"}}\par\vspace{0.5em}
        \textbf{Action:} {"action":"forward","value":40}\par\vspace{0.1em}
    \end{minipage}                                                                                                                                                                                                                                                                                                                                              \\
    \hline
    \begin{minipage}[t]{\linewidth}\centering
        \par\vspace{0.1em}
        \includegraphics[width=0.5\textwidth]{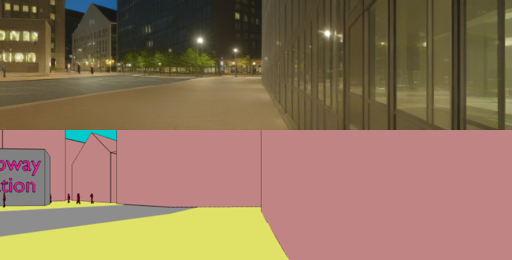}
    \end{minipage}\par\vspace{0.5em}

    \begin{minipage}[t]{\linewidth}
        \scriptsize
        \textbf{Log for step 15}\par\vspace{0.5em}
        \textbf{Task:} Navigate the urban environment to find a coffee shop before work.\par\vspace{0.5em}
        \textbf{Plan:} {"thought":"To efficiently reach the coffee shop, I should proceed forward towards the plaza, as it is likely to have coffee offerings among the commercial enterprises. The right side is blocked, making leftward or forward movement essential. Moving towards the center will allow me to explore the plaza without retracing steps unnecessarily."}\par\vspace{0.5em}
        \textbf{Memory:} {"previous\_observations":"I have observed a vibrant urban environment characterized by tall glass office buildings on the left, with clear sidewalks and established trees providing greenery and shade. The central area boasts an expansive plaza likely containing commercial enterprises, including cafes, illuminated by evening lights. The right side features mixed-use commercial buildings that add to the lively atmosphere, with retail options visible and inviting. Overall, the area is navigable, well-lit, and has undergone no significant changes since the last observation.","previous\_actions":"[step1: forward(40), step2: forward(30), step3: turnLeft(90)]"}\par\vspace{0.5em}
        \textbf{Warning:}    Warning! your can't move right.There is a/an building in your right in 1.16 m.\par\vspace{0.5em}
        \textbf{Observation:} {"observation":{"left":"Tall glass office buildings tower, reflecting the evening light, providing a modern aesthetic. There are clear sidewalks lined with established trees that offer some greenery and shade, creating a pleasant atmosphere.","front":"A spacious plaza dominates the central area, illuminated by warm lights. It likely houses various commercial enterprises, including cafes. The ground is well-paved, inviting for pedestrians and facilitating easy movement.","right":"Mixed-use commercial buildings with inviting retail options. The architecture is varied, adding character to the street. However, a building about 1.16 meters to the right blocks any potential movement in that direction. An existing parking area may also be noted.","compass":"there is no compass information"}}\par\vspace{0.5em}
        \textbf{Action:} {"action":"forward","value":30}\par\vspace{0.1em}
    \end{minipage}                                                                                                                                                                                                                                                                                                                            \\
    \hline
\end{longtable}

%% file: travelAgent.bbl
\begin{thebibliography}{66}
\providecommand{\natexlab}[1]{#1}
\providecommand{\url}[1]{\texttt{#1}}
\expandafter\ifx\csname urlstyle\endcsname\relax
  \providecommand{\doi}[1]{doi: #1}\else
  \providecommand{\doi}{doi: \begingroup \urlstyle{rm}\Url}\fi

\bibitem[Gehl(2010)]{gehl2010}
Jan Gehl.
\newblock \emph{Cities for People}.
\newblock Island Press, 2010.

\bibitem[Jacobs(1961)]{jacobs1961death}
Jane Jacobs.
\newblock \emph{The death and life of great American cities}.
\newblock Random House LLC, 1961.

\bibitem[Whyte(1980)]{whyte1980}
William~H. Whyte.
\newblock \emph{The Social Life of Small Urban Spaces}.
\newblock Project for Public Spaces, 1980.

\bibitem[Hillier and Hanson(1984)]{hillier1984}
Bill Hillier and Julienne Hanson.
\newblock \emph{The Social Logic of Space}.
\newblock Cambridge University Press, 1984.

\bibitem[Batty et~al.(2000)Batty, Chapman, Evans, Haklay, Kueppers, Shiode, Smith, and Torrens]{batty2000visualizing}
Michael Batty, David Chapman, Steve Evans, Mordechai Haklay, Stefan Kueppers, Naru Shiode, Andy Smith, and Paul~M Torrens.
\newblock Visualizing the city: communicating urban design to planners and decision-makers.
\newblock 2000.

\bibitem[Batty and Yang(2022)]{batty2022digital}
Michael Batty and Wei Yang.
\newblock \emph{A digital future for planning: Spatial planning reimagined}.
\newblock Digital Task Force for Planning, 2022.

\bibitem[Noyman(2022)]{noyman2022cityscope}
Ariel Noyman.
\newblock \emph{{{CityScope}} : An Urban Modeling and Simulation Platform}.
\newblock PhD thesis, Massachusetts Institute of Technology, Cambridge, Massachusetts, 2022.

\bibitem[Banerjee and Loukaitou-Sideris(2011)]{banerjee2011companion}
Tridib Banerjee and Anastasia Loukaitou-Sideris.
\newblock \emph{Companion to urban design}.
\newblock Routledge, 2011.

\bibitem[Perez et~al.(2016)Perez, Banos, and Pettit]{perez2016limitations}
P.~Perez, A.~Banos, and C.~Pettit.
\newblock Agent-based modelling for urban planning: Current limitations and future trends.
\newblock In \emph{Planning Support Science for Smarter Urban Futures}, pages 60--69. Springer, 2016.
\newblock \doi{10.1007/978-3-319-51957-9_4}.

\bibitem[Epstein(1999)]{epstein1999agent}
Joshua~M Epstein.
\newblock Agent-based computational models and generative social science.
\newblock \emph{Complexity}, 4\penalty0 (5):\penalty0 41--60, 1999.

\bibitem[Batty(2021)]{doi:10.1177/23998083211051139}
Michael Batty.
\newblock Multiple models.
\newblock \emph{Environment and Planning B: Urban Analytics and City Science}, 48\penalty0 (8):\penalty0 2129--2132, 2021.
\newblock ISSN 2399-8083, 2399-8091.
\newblock \doi{10.1177/23998083211051139}.

\bibitem[Chopra et~al.(2024)Chopra, Kumar, {Giray-Kuru}, Raskar, and {Quera-Bofarull}]{chopraLimitsAgencyAgentbased2024}
Ayush Chopra, Shashank Kumar, Nurullah {Giray-Kuru}, Ramesh Raskar, and Arnau {Quera-Bofarull}.
\newblock On the limits of agency in agent-based models, October 2024.

\bibitem[Luca et~al.(2021)Luca, Barlacchi, Lepri, and Pappalardo]{luca2021survey}
Massimiliano Luca, Gianni Barlacchi, Bruno Lepri, and Luca Pappalardo.
\newblock A survey on deep learning for human mobility.
\newblock \emph{ACM Computing Surveys (CSUR)}, 55\penalty0 (1):\penalty0 1--44, 2021.

\bibitem[Park et~al.(2023)Park, O'Brien, Cai, Morris, Liang, and Bernstein]{park2023generative}
Joon~Sung Park, Joseph O'Brien, Carrie~Jun Cai, Meredith~Ringel Morris, Percy Liang, and Michael~S Bernstein.
\newblock Generative agents: Interactive simulacra of human behavior.
\newblock In \emph{Proceedings of the 36th annual acm symposium on user interface software and technology}, pages 1--22, 2023.

\bibitem[Kaiya et~al.(2023)Kaiya, Naim, Kondic, Cortes, Ge, Luo, Yang, and Ahn]{kaiya2023lyfeagentsgenerativeagents}
Zhao Kaiya, Michelangelo Naim, Jovana Kondic, Manuel Cortes, Jiaxin Ge, Shuying Luo, Guangyu~Robert Yang, and Andrew Ahn.
\newblock Lyfe {{Agents}}: {{Generative}} agents for low-cost real-time social interactions, 2023.

\bibitem[Wei et~al.(2023)Wei, Wang, Schuurmans, Bosma, Ichter, Xia, Chi, Le, and Zhou]{weiChainofThoughtPromptingElicits2023}
Jason Wei, Xuezhi Wang, Dale Schuurmans, Maarten Bosma, Brian Ichter, Fei Xia, Ed~Chi, Quoc Le, and Denny Zhou.
\newblock Chain-of-{{Thought Prompting Elicits Reasoning}} in {{Large Language Models}}, January 2023.

\bibitem[Bonabeau(2002)]{bonabeauAgentbasedModelingMethods2002}
Eric Bonabeau.
\newblock Agent-based modeling: {{Methods}} and techniques for simulating human systems.
\newblock \emph{Proceedings of the National Academy of Sciences}, 99\penalty0 (suppl\_3):\penalty0 7280--7287, May 2002.
\newblock ISSN 0027-8424, 1091-6490.
\newblock \doi{10.1073/pnas.082080899}.

\bibitem[Chen(2012)]{CHEN2012166}
Liang Chen.
\newblock Agent-based modeling in urban and architectural research: {{A}} brief literature review.
\newblock \emph{Frontiers of Architectural Research}, 1\penalty0 (2):\penalty0 166--177, 2012.
\newblock ISSN 2095-2635.
\newblock \doi{10.1016/j.foar.2012.03.003}.

\bibitem[Lynch(1960)]{lynch1960image}
Kevin Lynch.
\newblock \emph{The image of the city}.
\newblock MIT press, 1960.

\bibitem[Ameli et~al.(2015)Ameli, Hamidi, Garfinkel-Castro, and Ewing]{ameli2015dobetter}
S.~H. Ameli, Shima Hamidi, Andrea Garfinkel-Castro, and R.~Ewing.
\newblock Do better urban design qualities lead to more walking in salt lake city, utah?
\newblock \emph{Journal of Urban Design}, 20:\penalty0 393 -- 410, 2015.
\newblock \doi{10.1080/13574809.2015.1041894}.

\bibitem[Billger et~al.(2017)Billger, Thuvander, and W{\"a}stberg]{billger2017search}
Monica Billger, Liane Thuvander, and Beata~Stahre W{\"a}stberg.
\newblock In search of visualization challenges: The development and implementation of visualization tools for supporting dialogue in urban planning processes.
\newblock \emph{Environment and Planning B: Urban Analytics and City Science}, 44\penalty0 (6):\penalty0 1012--1035, 2017.

\bibitem[Yin(2017)]{yin2017streetlevel}
L.~Yin.
\newblock Street level urban design qualities for walkability: Combining 2d and 3d gis measures.
\newblock \emph{Comput. Environ. Urban Syst.}, 64:\penalty0 288--296, 2017.
\newblock \doi{10.1016/j.compenvurbsys.2017.04.001}.

\bibitem[Shen et~al.(2018)Shen, Zeng, Ye, Arisona, Schubiger-Banz, Burkhard, and Qu]{shen2018streetvizor}
Qiaomu Shen, W.~Zeng, Y.~Ye, S.~Arisona, S.~Schubiger-Banz, R.~Burkhard, and Huamin Qu.
\newblock Streetvizor: Visual exploration of human-scale urban forms based on street views.
\newblock \emph{IEEE Transactions on Visualization and Computer Graphics}, 24:\penalty0 1004--1013, 2018.
\newblock \doi{10.1109/TVCG.2017.2744159}.

\bibitem[Smith et~al.(1998)Smith, Dodge, and Doyle]{smith1998visual}
Andy Smith, Martin Dodge, and Simon Doyle.
\newblock \emph{Visual communication in urban planning and urban design}.
\newblock University College London, Centre for Advanced Spatial Analysis (CASA), 1998.

\bibitem[Lin et~al.(2013)Lin, Lin, and Hu]{lin2013cognition}
Tianpeng Lin, Hui Lin, and Mingyuan Hu.
\newblock 3d visibility analysis in urban environment - cognition research based on vge.
\newblock \emph{ISPRS Annals of the Photogrammetry, Remote Sensing and Spatial Information Sciences}, pages 227--236, 2013.
\newblock \doi{10.5194/ISPRSANNALS-II-2-W1-227-2013}.

\bibitem[Tabrizian et~al.(2020)Tabrizian, Petrasova, Baran, Vukomanovic, Mitásová, and Meentemeyer]{tabrizian2020viewscape}
Payam Tabrizian, A.~Petrasova, Perver~K. Baran, Jelena Vukomanovic, H.~Mitásová, and R.~Meentemeyer.
\newblock High resolution viewscape modeling evaluated through immersive virtual environments.
\newblock \emph{ISPRS Int. J. Geo Inf.}, 9:\penalty0 445, 2020.
\newblock \doi{10.3390/ijgi9070445}.

\bibitem[Fonseca et~al.(2016)Fonseca, Valls, Redondo, and Villagrasa]{fonseca2016informal}
David Fonseca, Francesc Valls, Ernest Redondo, and Sergi Villagrasa.
\newblock Informal interactions in 3d education: Citizenship participation and assessment of virtual urban proposals.
\newblock \emph{Computers in Human Behavior}, 55:\penalty0 504--518, 2016.

\bibitem[Gath-Morad et~al.(2024)Gath-Morad, Gr{\"u}bel, Steemers, Sailer, Ben-Alon, H{\"o}lscher, and Aguilar]{gath2024role}
Michal Gath-Morad, Jascha Gr{\"u}bel, Koen Steemers, Kerstin Sailer, Lola Ben-Alon, Christoph H{\"o}lscher, and Leonel Aguilar.
\newblock The role of strategic visibility in shaping wayfinding behavior in multilevel buildings.
\newblock \emph{Scientific Reports}, 14\penalty0 (1):\penalty0 3735, 2024.

\bibitem[Montello(1993)]{montello1993scale}
Daniel~R Montello.
\newblock Scale and multiple psychologies of space.
\newblock \emph{Spatial information theory a theoretical basis for GIS}, pages 312--321, 1993.

\bibitem[Golledge(1999)]{golledge1999human}
Reginald~G Golledge.
\newblock Human wayfinding and cognitive maps.
\newblock \emph{Wayfinding behavior: Cognitive mapping and other spatial processes}, pages 5--45, 1999.

\bibitem[Javadi et~al.(2017)Javadi, Emo, Howard, and et~al.]{javadi2017hippocampal}
Amir-Homayoun Javadi, Beata Emo, Luke~R Howard, and et~al.
\newblock Hippocampal and prefrontal processing of network topology to simulate the future.
\newblock \emph{Nature Communications}, 8\penalty0 (1):\penalty0 1--11, 2017.
\newblock \doi{10.1038/s41467-017-00112-0}.

\bibitem[Epstein et~al.(2017)Epstein, Patai, Julian, and Spiers]{epstein2017cognitive}
Russell~A Epstein, Eva~Z Patai, Joshua~B Julian, and Hugo~J Spiers.
\newblock The cognitive map in humans: Spatial navigation and beyond.
\newblock \emph{Nature neuroscience}, 20\penalty0 (11):\penalty0 1504--1513, 2017.
\newblock \doi{10.1038/nn.4656}.

\bibitem[Burgess(2006)]{burgess2006spatial}
Neil Burgess.
\newblock Spatial memory: how egocentric and allocentric combine.
\newblock \emph{Trends in cognitive sciences}, 10\penalty0 (12):\penalty0 551--557, 2006.

\bibitem[Klatzky(2008)]{klatzky2008allocentric}
Roberta~L Klatzky.
\newblock Allocentric and egocentric spatial representations: Definitions, distinctions, and interconnections.
\newblock \emph{Spatial cognition}, pages 1--17, 2008.

\bibitem[Mou et~al.(2004)Mou, McNamara, Rump, and Xiao]{mou2004roles}
Weimin Mou, Timothy~P McNamara, Bj{"o}rn Rump, and Chengli Xiao.
\newblock The roles of egocentric and allocentric spatial representations in locomotion and reorientation.
\newblock \emph{Journal of Experimental Psychology: Learning, Memory, and Cognition}, 30\penalty0 (6):\penalty0 1322, 2004.

\bibitem[Burgess et~al.(2017)Burgess, Epstein, and Spiers]{burgess2017cognitive}
Neil Burgess, Russell~A Epstein, and Hugo~J Spiers.
\newblock The cognitive map in humans: spatial navigation and beyond.
\newblock \emph{Nature neuroscience}, 20\penalty0 (11):\penalty0 1504--1513, 2017.

\bibitem[Milivojevic and Doeller(2013)]{milivojevic2013mnemonic}
Branka Milivojevic and Christian~F Doeller.
\newblock Mnemonic networks in the hippocampal formation: From spatial maps to temporal and conceptual codes.
\newblock \emph{Journal of Experimental Psychology: General}, 142\penalty0 (4):\penalty0 1231, 2013.

\bibitem[Bankes(2002)]{bankesAgentbasedModelingRevolution2002}
Steven~C. Bankes.
\newblock Agent-based modeling: {{A}} revolution?
\newblock \emph{Proceedings of the National Academy of Sciences}, 99\penalty0 (suppl\_3):\penalty0 7199--7200, May 2002.
\newblock ISSN 0027-8424, 1091-6490.
\newblock \doi{10.1073/pnas.072081299}.

\bibitem[Nguyen et~al.(2021)Nguyen, Powers, Urquhart, Farrenkopf, and Guckert]{nguyenOverviewAgentbasedTraffic2021}
Johannes Nguyen, Simon~T. Powers, Neil Urquhart, Thomas Farrenkopf, and Michael Guckert.
\newblock An overview of agent-based traffic simulators.
\newblock \emph{Transportation Research Interdisciplinary Perspectives}, 12:\penalty0 100486, December 2021.
\newblock ISSN 25901982.
\newblock \doi{10.1016/j.trip.2021.100486}.

\bibitem[Grignard et~al.(2018)Grignard, Maci{\`a}, Pastor, Noyman, Zhang, and Larson]{grignard2018cityscope}
Arnaud Grignard, N{\'u}ria Maci{\`a}, Luis~Alonso Pastor, Ariel Noyman, Yan Zhang, and Kent Larson.
\newblock Cityscope andorra: A multi-level interactive and tangible agent-based visualization.
\newblock In \emph{AAMAS}, pages 1939--1940, 2018.

\bibitem[Filomena and Verstegen(2021)]{filomenaModellingEffectLandmarks2021}
Gabriele Filomena and Judith~A. Verstegen.
\newblock Modelling the effect of landmarks on pedestrian dynamics in urban environments.
\newblock \emph{Computers, Environment and Urban Systems}, 86:\penalty0 101573, March 2021.
\newblock ISSN 01989715.
\newblock \doi{10.1016/j.compenvurbsys.2020.101573}.

\bibitem[Railsback and Grimm(2019)]{railsback2019agent}
Steven~F Railsback and Volker Grimm.
\newblock \emph{Agent-based and individual-based modeling: a practical introduction}.
\newblock Princeton university press, 2019.

\bibitem[Zhang et~al.(2024)Zhang, Chen, Sheng, Wang, and Chua]{zhang2024generative}
An~Zhang, Yuxin Chen, Leheng Sheng, Xiang Wang, and Tat-Seng Chua.
\newblock On generative agents in recommendation.
\newblock In \emph{Proceedings of the 47th international ACM SIGIR conference on research and development in Information Retrieval}, pages 1807--1817, 2024.

\bibitem[Kobayashi et~al.(2023)Kobayashi, Takeda, Yamazaki, and Kamisaka]{kobayashi2023modeling}
Akihiro Kobayashi, Naoto Takeda, Yudai Yamazaki, and Daisuke Kamisaka.
\newblock Modeling and generating human mobility trajectories using transformer with day encoding.
\newblock In \emph{Proceedings of the 1st International Workshop on the Human Mobility Prediction Challenge}, pages 7--10, 2023.

\bibitem[Li et~al.(2020)Li, Li, Chen, Chen, and Li]{li2020agent}
Feixue Li, Zhifeng Li, Honghua Chen, Zhenjie Chen, and Manchun Li.
\newblock An agent-based learning-embedded model (abm-learning) for urban land use planning: A case study of residential land growth simulation in shenzhen, china.
\newblock \emph{Land Use Policy}, 95:\penalty0 104620, 2020.

\bibitem[Chen et~al.(2019)Chen, Suhr, Misra, Snavely, and Artzi]{chen2019touchdown}
Howard Chen, Alane Suhr, Dipendra Misra, Noah Snavely, and Yoav Artzi.
\newblock Touchdown: Natural language navigation and spatial reasoning in visual street environments.
\newblock In \emph{Proceedings of the IEEE/CVF Conference on Computer Vision and Pattern Recognition}, pages 12538--12547, 2019.

\bibitem[Reid et~al.(2024)Reid, Savinov, Teplyashin, Lepikhin, Lillicrap, Alayrac, Soricut, Lazaridou, Firat, Schrittwieser, et~al.]{reid2024gemini}
Machel Reid, Nikolay Savinov, Denis Teplyashin, Dmitry Lepikhin, Timothy Lillicrap, Jean-baptiste Alayrac, Radu Soricut, Angeliki Lazaridou, Orhan Firat, Julian Schrittwieser, et~al.
\newblock Gemini 1.5: {{Unlocking}} multimodal understanding across millions of tokens of context.
\newblock \emph{arXiv preprint arXiv:2403.05530}, 2024.

\bibitem[Yang et~al.(2025)Yang, Ding, Brown, Qi, and Xie]{yang2025v}
Jihan Yang, Runyu Ding, Ellis Brown, Xiaojuan Qi, and Saining Xie.
\newblock V-irl: Grounding virtual intelligence in real life.
\newblock In \emph{European Conference on Computer Vision}, pages 36--55. Springer, 2025.

\bibitem[Huang et~al.(2024)Huang, Liu, Chen, Wang, Wang, Lian, Wang, Tang, and Chen]{huangUnderstandingPlanningLLM2024}
Xu~Huang, Weiwen Liu, Xiaolong Chen, Xingmei Wang, Hao Wang, Defu Lian, Yasheng Wang, Ruiming Tang, and Enhong Chen.
\newblock Understanding the planning of {{LLM}} agents: {{A}} survey, February 2024.

\bibitem[Yao et~al.(2023)Yao, Zhao, Yu, Du, Shafran, Narasimhan, and Cao]{yaoReActSynergizingReasoning2023}
Shunyu Yao, Jeffrey Zhao, Dian Yu, Nan Du, Izhak Shafran, Karthik Narasimhan, and Yuan Cao.
\newblock {{ReAct}}: {{Synergizing Reasoning}} and {{Acting}} in {{Language Models}}, March 2023.

\bibitem[Yao et~al.()Yao, Yu, Zhao, Shafran, Griffiths, Cao, and Narasimhan]{yaoTreeThoughtsDeliberate}
Shunyu Yao, Dian Yu, Jeffrey Zhao, Izhak Shafran, Thomas~L Griffiths, Yuan Cao, and Karthik Narasimhan.
\newblock Tree of {{Thoughts}}: {{Deliberate Problem Solving}} with {{Large Language Models}}.

\bibitem[Ghaffarzadegan et~al.(2023)Ghaffarzadegan, Majumdar, Williams, and Hosseinichimeh]{ghaffarzadegan2023generative}
Navid Ghaffarzadegan, Aritra Majumdar, Ross Williams, and Niyousha Hosseinichimeh.
\newblock Generative agent-based modeling: Unveiling social system dynamics through coupling mechanistic models with generative artificial intelligence.
\newblock \emph{arXiv preprint arXiv:2309.11456}, 2023.

\bibitem[Hatalis et~al.(2024)Hatalis, Christou, Myers, Jones, Lambert, {Amos-Binks}, Dannenhauer, and Dannenhauer]{hatalisMemoryMattersNeed2024}
Kostas Hatalis, Despina Christou, Joshua Myers, Steven Jones, Keith Lambert, Adam {Amos-Binks}, Zohreh Dannenhauer, and Dustin Dannenhauer.
\newblock Memory {{Matters}}: {{The Need}} to {{Improve Long-Term Memory}} in {{LLM-Agents}}.
\newblock \emph{Proceedings of the AAAI Symposium Series}, 2\penalty0 (1):\penalty0 277--280, January 2024.
\newblock ISSN 2994-4317.
\newblock \doi{10.1609/aaaiss.v2i1.27688}.

\bibitem[Mirzadeh et~al.(2024)Mirzadeh, Alizadeh, Shahrokhi, Tuzel, Bengio, and Farajtabar]{mirzadehGSMSymbolicUnderstandingLimitations2024}
Iman Mirzadeh, Keivan Alizadeh, Hooman Shahrokhi, Oncel Tuzel, Samy Bengio, and Mehrdad Farajtabar.
\newblock {{GSM-Symbolic}}: {{Understanding}} the {{Limitations}} of {{Mathematical Reasoning}} in {{Large Language Models}}, October 2024.

\bibitem[Podell et~al.(2023)Podell, English, Lacey, Blattmann, Dockhorn, M{\"u}ller, Penna, and Rombach]{podell2023sdxl}
Dustin Podell, Zion English, Kyle Lacey, Andreas Blattmann, Tim Dockhorn, Jonas M{\"u}ller, Joe Penna, and Robin Rombach.
\newblock Sdxl: Improving latent diffusion models for high-resolution image synthesis.
\newblock \emph{arXiv preprint arXiv:2307.01952}, 2023.

\bibitem[Redmon et~al.(2015)Redmon, Divvala, Girshick, and Farhadi]{redmon2015you}
Joseph Redmon, Santosh Divvala, Ross Girshick, and Ali Farhadi.
\newblock You only look once: unified, real-time object detection (2015).
\newblock \emph{arXiv preprint arXiv:1506.02640}, 825, 2015.

\bibitem[Achiam et~al.(2023)Achiam, Adler, Agarwal, Ahmad, Akkaya, Aleman, Almeida, Altenschmidt, Altman, Anadkat, et~al.]{achiam2023gpt}
Josh Achiam, Steven Adler, Sandhini Agarwal, Lama Ahmad, Ilge Akkaya, Florencia~Leoni Aleman, Diogo Almeida, Janko Altenschmidt, Sam Altman, Shyamal Anadkat, et~al.
\newblock Gpt-4 technical report.
\newblock \emph{arXiv preprint arXiv:2303.08774}, 2023.

\bibitem[Cheng et~al.(2021)Cheng, Misra, Schwing, Kirillov, and Girdhar]{cheng2021masked}
B~Cheng, I~Misra, AG~Schwing, A~Kirillov, and R~Girdhar.
\newblock Masked-attention mask transformer for universal image segmentation. arxiv 2022.
\newblock \emph{arXiv preprint arXiv:2112.01527}, 2021.

\bibitem[Kim(1999)]{kim1999spatial}
Young~Ook Kim.
\newblock \emph{Spatial Configuration, Spatial Cognition and Spatial Behaviour: the role of architectural intelligibility in shaping spatial experience}.
\newblock University of London, University College London (United Kingdom), 1999.

\bibitem[Kantardzic(2011)]{kantardzic2011data}
Mehmed Kantardzic.
\newblock \emph{Data mining: concepts, models, methods, and algorithms}.
\newblock John Wiley \& Sons, 2011.

\bibitem[Petukhova et~al.(2024)Petukhova, Matos-Carvalho, and Fachada]{petukhova2024text}
Alina Petukhova, Jo{\~a}o~P Matos-Carvalho, and Nuno Fachada.
\newblock Text clustering with large language model embeddings.
\newblock \emph{International Journal of Cognitive Computing in Engineering}, 2024.

\bibitem[Blei et~al.(2003)Blei, Ng, and Jordan]{blei2003latent}
David~M Blei, Andrew~Y Ng, and Michael~I Jordan.
\newblock Latent dirichlet allocation.
\newblock \emph{Journal of machine Learning research}, 3\penalty0 (Jan):\penalty0 993--1022, 2003.

\bibitem[Hutto and Gilbert(2014)]{hutto2014vader}
Clayton Hutto and Eric Gilbert.
\newblock Vader: A parsimonious rule-based model for sentiment analysis of social media text.
\newblock In \emph{Proceedings of the international AAAI conference on web and social media}, volume~8, pages 216--225, 2014.

\bibitem[Ferrara(2023)]{ferrara2023should}
Emilio Ferrara.
\newblock Should chatgpt be biased? challenges and risks of bias in large language models.
\newblock \emph{arXiv preprint arXiv:2304.03738}, 2023.

\bibitem[Kotek et~al.(2023)Kotek, Dockum, and Sun]{kotek2023gender}
Hadas Kotek, Rikker Dockum, and David Sun.
\newblock Gender bias and stereotypes in large language models.
\newblock In \emph{Proceedings of the ACM collective intelligence conference}, pages 12--24, 2023.

\bibitem[Manvi et~al.(2024)Manvi, Khanna, Burke, Lobell, and Ermon]{manvi2024large}
Rohin Manvi, Samar Khanna, Marshall Burke, David Lobell, and Stefano Ermon.
\newblock Large language models are geographically biased.
\newblock \emph{arXiv preprint arXiv:2402.02680}, 2024.

\end{thebibliography}
